\documentclass[runningheads]{llncs}

\usepackage{accv}

\usepackage{accvabbrv}

\usepackage{graphicx}
\usepackage{booktabs}
\usepackage{makecell}
\usepackage{multirow}
\usepackage{bbding}

\usepackage[accsupp]{axessibility}  

\usepackage{hyperref}

\usepackage{orcidlink}

\begin{document}

\title{Structure-Centric Robust Monocular Depth Estimation via Knowledge Distillation} 

\titlerunning{Scent-Depth: Robost MDE via KD}

\author{Runze Chen\inst{1,2}\orcidlink{0000-0002-6599-7898} \and
Haiyong Luo\inst{2}\textsuperscript{$\dagger$}\orcidlink{0000-0001-6827-4225} \and
Fang Zhao\inst{1}\textsuperscript{$\dagger$}\orcidlink{0000-0002-4784-5778}\and Jingze Yu\inst{3}\orcidlink{0009-0008-5656-985X} \and Yupeng Jia\inst{1,2}\orcidlink{0009-0001-2654-5629} \and Juan Wang\inst{4} \and Xuepeng Ma\inst{4}}

\begingroup
\renewcommand\thefootnote{$\dagger$}
\footnotetext{Joint corresponding authors: Haiyong Luo and Fang Zhao.}
\endgroup

\authorrunning{R.~Chen et al.}

\institute{Beijing University of Posts and Telecommunications, Beijing, China  \and Institute of Computing Technology Chinese Academy of Sciences, Beijing, China  \and Guangdong Hong Kong Macao Greater Bay Area National Technology Innovation Center, Guangdong, China \and 
Shouguang Cheng Zhi Feng Xing Technology Co., Ltd, Shandong, China \email{yhluo@ict.ac.cn},~\email{zfsse@bupt.edu.cn}}

\maketitle

\begin{abstract}
  Monocular depth estimation, enabled by self-supervised learning, is a key technique for 3D perception in computer vision. However, it faces significant challenges in real-world scenarios, which encompass adverse weather variations, motion blur, as well as scenes with poor lighting conditions at night. Our research reveals that we can divide monocular depth estimation into three sub-problems: depth structure consistency, local texture disambiguation, and semantic-structural correlation. Our approach tackles the non-robustness of existing self-supervised monocular depth estimation models to interference textures by adopting a structure-centered perspective and utilizing the scene structure characteristics demonstrated by semantics and illumination. We devise a novel approach to reduce over-reliance on local textures, enhancing robustness against missing or interfering patterns. Additionally, we incorporate a semantic expert model as the teacher and construct inter-model feature dependencies via learnable isomorphic graphs to enable aggregation of semantic structural knowledge. Our approach achieves state-of-the-art out-of-distribution monocular depth estimation performance across a range of public adverse scenario datasets. It demonstrates notable scalability and compatibility, without necessitating extensive model engineering. This showcases the potential for customizing models for diverse industrial applications. The source code for our method is available at \href{https://github.com/chenrz925/Scent-Depth}{Scent-Depth}.
\end{abstract}

\section{Introduction}
\label{sec:intro}

\begin{figure*}[htbp]
    \centering
    \includegraphics[width=0.8\textwidth]{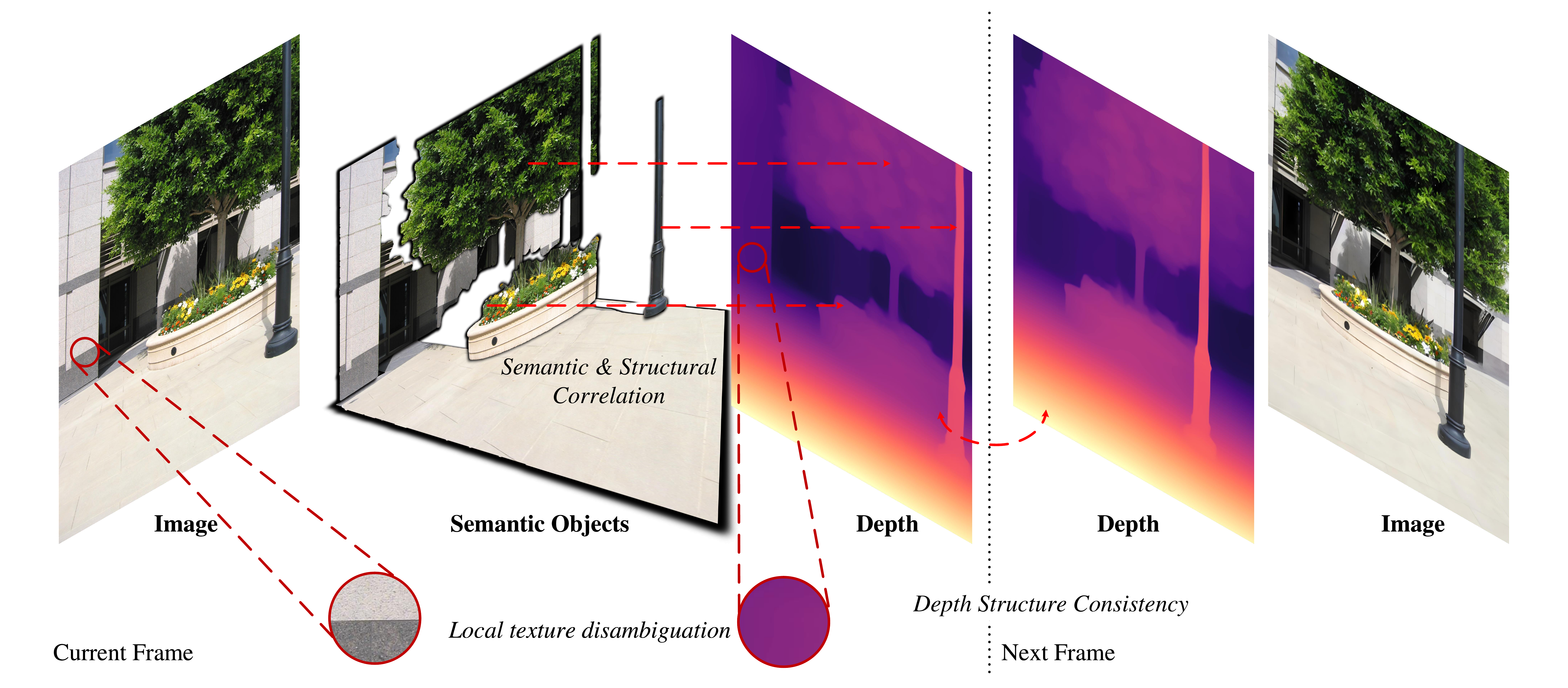}
    \caption{Key Sub-Problems in Monocular Depth Estimation. This figure highlights three areas critical for enhancing depth estimation: \textit{Depth Structure Consistency}, ensuring smooth depth transitions across frames; \textit{Local Texture Disambiguation}, addressing the challenge of the model's over-dependence on local textures that compromises depth estimation robustness, by improving performance in diverse or texture-sparse environments; and \textit{Semantic and Structural Correlation}, leveraging object semantics and structure for continuous depth inference.}
    \label{fig:sub-problems}
\end{figure*}

Monocular depth estimation (MDE) is a popular research topic in the field of computer vision due to its broad applications in fields such as autonomous driving, robotics, and augmented reality. However, accurately estimating objects' depth from a single 2D image remains a challenging task despite significant research efforts. Synthesizing new perspectives across multiple frames enables training a monocular affine-invariant depth estimation model in a self-supervised manner \cite{DBLP:conf/cvpr/ZhouBSL17}. This approach imposes high demands on the image quality of adjacent frames, as even minor device or environmental disturbances can significantly degrade training quality \cite{DBLP:conf/iccv/GodardAFB19}.

Thus, it is crucial to investigate methods for enhancing the robustness of depth estimation models built through self-supervision based on clear visual sequences, particularly in adverse environments. Moreover, models that solely rely on pixel-level features fail to capture the critical structural information of objects, which negatively impacts their performance in complex and noisy real-world environments. The current approach constrains the model from learning reasonable affine-invariant depth through direct projection between pixels, reflecting the consistency of scene structure during motion. However, not all textures serve to represent scene structure effectively for the model. Over-reliance on structure-agnostic textures may lead to significant performance degradation in the presence of external disturbances, such as occlusion, adverse weather, equipment malfunction, and inconsistent lighting conditions \cite{DBLP:conf/nips/KongXHNCO23}. Therefore, effectively incorporating structural information into the model becomes crucial in enhancing its depth estimation performance in various practical scenarios.

To distinguish the model's understanding of effective scene structure, we propose to decouple the problem of representing structural information for robust monocular depth estimation into three sub-problems as displayed in Figure \ref{fig:sub-problems}:
\begin{itemize}
    \item \textit{Depth structure consistency}, which ensures that the relative depth structure of the scene remains consistent across adjacent frames during camera motion, avoiding abrupt changes in estimated depths. 
    \item \textit{Local texture disambiguation} refers to resolving degradation issues in depth estimation models caused by overfitting of local detail textures, which can lead to estimation ambiguities in adverse scenarios due to interfering textures.
    \item \textit{Semantic and structural correlation} illustrates the inherent link between an object's semantics (e.g., shape, function) and its visual structure (e.g., size, position, orientation), where fixed semantic properties guide the estimation of continuous depth across different views.
\end{itemize}

Based on the above considerations, we propose solutions for each of the aforementioned sub-problems as illustrated in Figure \ref{fig:structure-centric-mde}. Adopting the retinex principle, we decouple ambient illumination from object surface reflectance in images. This reduces the monocular depth estimation model’s over-reliance on local textures of scene objects, enabling robustness against missing textures and distracting patterns. Further, we construct an isomorphic graph between the depth model and a semantic expert model \cite{kirillov2023segment} in a learnable manner. This achieves distillation of structural knowledge from the semantic expert into the monocular depth model. By employing the aforementioned methods, we substantially enhance the robustness of monocular depth estimation models trained on clear datasets in challenging scenarios, such as adverse weather variations, motion blur, and scenes with poor lighting conditions at night. This explicitly builds structure-centric monocular depth estimation.

In conclusion, our contributions are four-fold:

\begin{itemize}
    \item We propose decoupling monocular depth estimation into three sub-problems: local texture disambiguation, semantic-structural correlation, and depth structure consistency. These highlight key challenges for monocular depth estimation methods in complex environments.
    \item We decouple ambient illumination from object surface reflectance to reduce the depth estimation model's strong reliance on local textures. This mitigates performance degradation stemming from over-dependence on local patterns under adverse weather, intense motion, and poor lighting conditions.
    \item Our designed isomorphic graph distillation enables the aggregation of features across different tasks. It transforms inter-channel dependencies into a learnable loss, thereby distilling semantic structural knowledge into the depth model. This effectively learns mappings between teacher and student feature spaces, enabling effective knowledge transfer.
    \item This paper presents an effective benchmark for robustness and generalization under common real-world conditions. It reveals the challenges faced by current state-of-the-art models under these scenarios. It also demonstrates that our method is effective, achieving state-of-the-art performance across these challenging scenarios.
\end{itemize}

In this paper, we present a novel approach for monocular depth estimation that breaks down the problem into individual sub-problems. We address those sub-problems through the development of specific deep-learning frameworks designed to meet the demands of complex real-world scenes. Moving forward, we describe the pipeline of our proposed method, present experimental results that validate its effectiveness, and provide our conclusions. Our method yields state-of-the-art performance on widely-used public benchmark datasets, even under various forms of corruption. This demonstrates the method's strong potential for real-world applications in complex scenes.

\section{Related Works}
\label{sec:rel-works}

\subsection{Self-supervised Monocular Depth Estimation}

The technique of using multi-view synthesis and pose estimation as proxy tasks proves to be an effective method for accomplishing self-supervised training of affine invariant depth estimation models. Employing a warping-based view synthesis approach with monocular image sequences \cite{DBLP:conf/cvpr/ZhouBSL17} and stereo images \cite{DBLP:conf/cvpr/GodardAB17} enables the model to learn from rudimentary 3D object structural details, providing an elegant solution for monocular depth estimation. The scale-invariant nature of object structure is a crucial aspect of its defining features. Furthermore, employing a multi-scale sampling approach \cite{DBLP:conf/iccv/GodardAFB19} in depth estimation provides a strategy to counteract the restrictive impact of local texture on the model's capacity to perceive the underlying structure of objects. In the context of image sequences, it is reasonable to posit that there exists a relative consistency in the structural organization of depth maps across different frames \cite{DBLP:conf/nips/BianLWZSC019}. The fusion of Vision Transformers (ViT) and convolutional features \cite{DBLP:conf/3dim/ZhaoZPTGZHTM22,DBLP:conf/cvpr/NingFGN23} effectively improves the modeling capacity for long-range structural features. Existing robust self-supervised monocular MDE works \cite{DBLP:conf/iccv/SaundersVM23,DBLP:conf/iccv/GasperiniMJNT23,DBLP:conf/iccv/WangZY0X0021} mainly focus on learning models from huge datasets containing various scenes. Training a robust MDE model using clear, interference-free data is a significant challenge, as it must be capable of handling real-world conditions such as rain, snow, fog, motion blur, and night-time scenarios effectively.

\subsection{Knowledge-Enhanced Monocular Depth Estimation}

An effective approach to address challenges in the real world involves introducing more knowledge guidance during the self-supervised process, serving as an enhancement for self-supervised depth estimation training. Knowledge distillation is an effective approach, and many works \cite{10306272,10322131} accomplish the compression of the monocular depth estimation model through this approach. Constructing pseudo depth maps through supervised deep estimation models \cite{DBLP:conf/iccv/RanftlBK21} to guide distillation in self-supervised training has to some extent enhanced the training effectiveness of self-supervised depth estimation models \cite{DBLP:journals/ijcv/BianZWLZSCR21}. This approach essentially introduces implicit supervised information into self-supervised training. Semantic labels of objects can robustly represent the structure and contour of objects in the scene \cite{DBLP:conf/eccv/KlingnerTMF20}. However, in common video sequence acquisition scenarios, the annotation of semantic labels typically consumes a substantial amount of manual effort and incurs high costs. Additional sensors, such as inertial sensors, can also assist in the scale consistency training of depth estimation \cite{10.1007/978-3-031-19839-7_9}. Integrating reliance on inertial sensors to some extent reduces the model's usability for devices lacking inertial sensors. The use of affine invariant depth maps generated by unsupervised depth estimation models for dense reconstruction to create mesh grids, and the distillation of depth estimation models through the synthesis of pseudo depth maps from these mesh grids can provide unsupervised depth estimation of a higher surface quality \cite{10377594}. The multi-stage training and its dense reconstruction process also imply more computational costs during the training process. Introducing additional knowledge requires consideration of various factors, including data diversity and training costs. Simultaneously, we believe robustness in adverse scenarios is crucial.

\subsection{Structure-Centric Knowledge in Adverse Scenarios}

Self-supervised monocular depth estimation shows promising prospects in affine invariant depth estimation and relative pose prediction. However, in real-world application scenarios, visual degradation caused by weather, lighting conditions, and sensor malfunctions poses a notable challenge \cite{MING202114}. Maintaining the robustness of neural networks towards visually degraded scenarios is always a focus in the academic field \cite{DBLP:conf/iclr/HendrycksD19}. In self-supervising depth estimation, the out-of-distribution generalization ability for adverse scenarios \cite{DBLP:conf/nips/KongXHNCO23} determines whether the depth estimation model can adapt to wide applications in the real world. How to generalize the monocular depth estimation model trained on clear visual sequence data to a wider and more complex application scenario still needs more attention. The semantic information of objects has consistency with the structure of the scene \cite{9578162}. Guiding through the semantic prediction of the teacher model can bring significant performance improvement \cite{DBLP:conf/iclr/GuiziliniHLAG20} in monocular depth estimation. The introduction of training constraints through semantic labels is also proven to enhance the domain transfer ability of depth estimation models \cite{Lopez-Rodriguez2023}. Illumination in an image represents the interaction of the object structure in the scene with the natural light source \cite{DBLP:conf/cvpr/RematasF20}. Unlike the interference texture appearing in the image, the scene structure characteristics shown by the illumination remain unaffected in adverse weather and various light intensities \cite{GOYAL2024102151}. Existing works \cite{Chen2023} also pay attention to the connection between depth structure and illumination, guiding the learning of illumination through depth maps to provide low-light enhancement algorithms. Hence, we aim to enhance the out-of-distribution generalization ability of monocular depth estimation models in challenging scenarios never encountered before, guided by the structural semantic features of mature semantic expert models and the scene structure characteristics represented by illumination.

\begin{figure}
    \centering
    \includegraphics[width=0.8\textwidth]{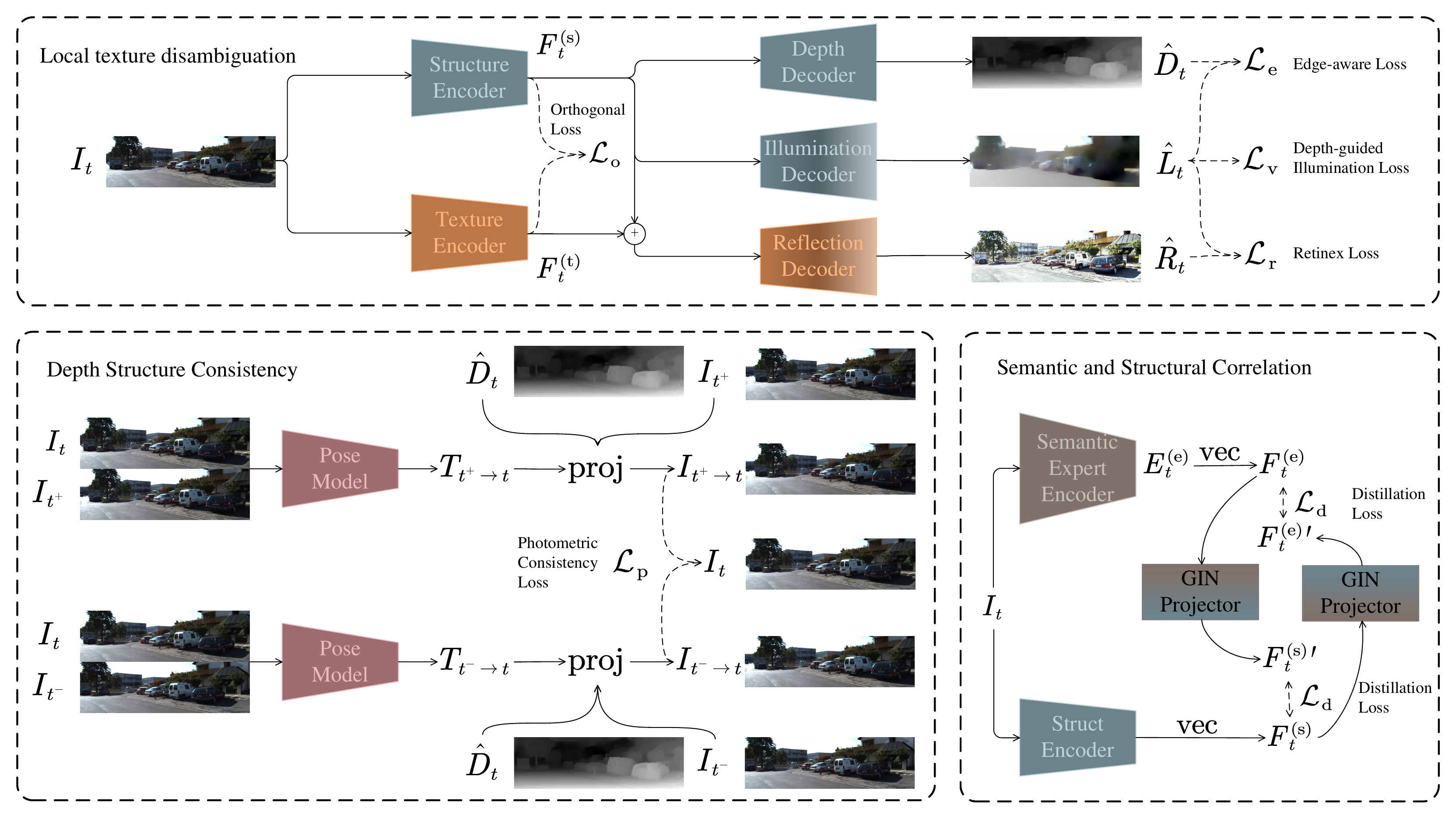}
    \caption{Overview of the Structure-Centric Monocular Depth Estimation Approach. The solid lines in the figure indicate data flow, while the dashed lines represent constraints during the optimization process. The operation \textrm{vec} represents the process of flattening these features into vectors. The operation \textrm{proj} denotes the re-projection of views.}
    \label{fig:structure-centric-mde}
\end{figure}

\section{Structure-Centric Depth Estimation}
\label{sec:scent}

We describe how our proposed structure-centric monocular depth estimation tackles the degradation problem under common camera corruptions in real-world scenarios. Specifically, guided by the three sub-problems including depth-structure consistency, local texture disambiguation, and semantic-structural correlation, we demonstrate how our method mitigates performance degradation due to interference from adverse weather, motion, and poor lighting conditions at night. We accomplish this through knowledge distillation techniques to realize robust structure-centric monocular depth estimation.

\subsection{Depth Structure Consistency}


Given a monocular image sequence $I=\left\{\cdots,I_t\in\mathbb{R}^{3\times W\times H},\cdots\right\}$ with camera intrinsics determined by $K$, the objective of monocular estimation is to establish a deep learning model for accurate estimation of the corresponding depth $\hat{D}_t\in\mathbb{R}^{1\times W\times H}$ from monocular image frame $I_t$. Given the difficulty and high cost of measuring depth ground truth $D_t$, we typically rely on unsupervised learning methods. This requires our monocular depth estimation approach to extract more sufficient scene structural information from a single image.

Monocular depth estimation typically involves synthesizing the view $I_{t^\prime\rightarrow t}$ by using the estimated relative pose $\hat{T}_{t\rightarrow t^\prime}$ and the estimated depth map $\hat{D}_t$, with respect to the source frame $I_{t^\prime}$ and the target frame $I_{t}$. We can express the synthesis operation as
\begin{equation}
    I_{{t^\prime\rightarrow t}}=I_{t^\prime}\left\langle\mathrm{proj}\left(\hat{D}_t,\hat{T}_{t\rightarrow t^{\prime}}, K\right)\right\rangle,
\end{equation}
where $\mathrm{proj}(\cdot)$ produces the two-dimensional positions resulting from the projection of depths $D_t$ onto the image $I_{t^\prime}$, $\left\langle\cdot\right\rangle$ upsamples the estimation to match the shape of $I_{t^\prime}$, and $I_{{t^\prime\rightarrow t}}$ is an approximation of $I_t$ obtained by projecting $I_{t^\prime}$. Monocular depth estimation aims to model depth structure consistency, which leverages the consistency of depth structure between adjacent frames to accomplish view synthesis tasks. Drawing on \cite{DBLP:conf/cvpr/ZhouBSL17,DBLP:journals/tci/ZhaoGFK17}, we utilize $\mathcal{L}_\mathrm{p}$  to impose constraints on the quality of re-projected views,
\begin{equation}
    \begin{array}{c}
        \mathcal{L}_\mathrm{p}^{u,v}(I_t,I_{t^\prime\rightarrow t})=\frac{\alpha}{2}\left(1-\mathrm{ssim}\left(I_t,I_{t^\prime\rightarrow t}\right)\right)+(1-\alpha)\left\Vert I_t-I_{t^\prime\rightarrow t}\right\Vert _1\\
        \mathcal{L}_\mathrm{p}=\sum\mu\mathcal{L}_\mathrm{p}(I_t,I_{t^\prime\rightarrow t})
    \end{array},
\end{equation}
where $\mu$ is a dynamic mask \cite{DBLP:conf/iccv/GodardAFB19}, and $\mathrm{ssim}$ measures the luminance consistency between two image frames \cite{DBLP:journals/tci/ZhaoGFK17}.	

The static environmental structure in the scene effectively provides structural information for depth estimation in the image sequence, which means we need to pay attention to the consistency of scene structure between adjacent frames. While maintaining the consistency between the synthesized view $I_{t^\prime\rightarrow t}$ and the target frame $I_{t^\prime}$, the estimated depth $\hat{D}_{t^\prime}$ of the source frame also exhibits structural consistency with the estimated depth $\hat{D}_t$ of the target frame. 

\subsection{Local Texture Disambiguation}

Monocular depth estimation infers scene depth from camera motion cues. The shading patterns induced by ambient light reveal structural features of objects. Such light-dark contrasts in scenes facilitate important clues about scene structures that are critical for depth models. However, based on color constancy, abundant surface textures of objects appear unaffected by scene structures in camera-captured images. Conversely, excessively complex textures may impede depth judgment, necessitating the extraction of valid cues. By adopting the Retinex principle, we decouple structural information and surface textures in scenes. This reduces model sensitivity to surface patterns under complex imaging conditions and lowers susceptibility to noise interference. 

For our camera observation $I_t$, there exist corresponding reflectance $R_t$ and illumination $L_t$ terms that satisfy the relation $I_t=R_t*L_t$. The reflectance term $R_t$ primarily exhibits the surface texture patterns of objects in the scene, while the illumination term $L_t$ mainly captures the ambient shading on structural surfaces within the scene. We leverage the estimated depth $\hat{D}_t$ along with $I_t$ to jointly optimize the illumination estimation $\hat{L}_t$. The optimization loss is
\begin{equation}
\begin{split}
    \mathcal{L}_\mathrm{v}=\sum\max\left(\left\|I_t-\hat{L}_t\right\|,\alpha\right)+\sum\left\|\nabla\hat{L}_t\right\|\cdot\mathrm{e}^{-\left\|\nabla\hat{D}_t\right\|-\left\|\nabla I_t\right\|},
\end{split}
\end{equation}
where the gradient of $\hat{D}_t$ is not recorded during training. We apply a suitable threshold of $\alpha=0.2$ in $\mathcal{L}_\mathrm{v}$ to preclude absolute proximity between local pixels. This ensures relatively consistent luminance within regions, which facilitates reducing the influence of localized textures.

We leverage the complementary nature of ambient illumination and scene depth. Specifically, we utilize an edge-aware loss $\mathcal{L}_\mathrm{e}$ between the illumination component of ambient light and the depth map,
\begin{equation}
    \mathcal{L}_\mathrm{e}=\sum{\left\Vert\nabla\hat{D}_t\right\Vert}\cdot\mathbf{e}^{-\left\Vert\nabla \hat{L}_t\right\Vert},
\end{equation}
where the gradient of $\hat{L}_t$ is not recorded during training. The same encoder extracts features to generate the illumination and depth map. We guide the illumination estimation using the gradients of the depth map and original image. This encourages continuous illumination in regions with contiguous surface structures.

We derive the reflectance $\hat{R}_t$ from the illumination $\hat{L}_t$. As reflectance mostly captures fine surface textures, we aim to reduce model sensitivity to superficial details without impairing the extraction of structural textures. We optimize the model's computation of the reflectance $\hat{R}_t$ through the loss $\mathcal{L}_\mathrm{r}$,
\begin{equation}
    \mathcal{L}_\mathrm{r}=\mathrm{ssim}\left(\hat{R}_t*\hat{L}_t,I_t\right),
\end{equation}
where $\hat{L}_r$ does not participate in gradient calculation during training. We desire the reflectance model to adequately extract fine surface textures across all objects in the scene structure while minimizing the proportion of structural cues. 

In Figure \ref{fig:structure-centric-mde}, the structure encoder generates the feature embedding $E_t^\mathrm{(s)}\in\mathbb{R}^{C\times H\times W}$, while the texture encoder produces $E_t^\mathrm{(t)}\in\mathbb{R}^{C\times H\times W}$. We expect $E_t^\mathrm{(s)}$ to represent all structural cues in the scene, and $E_t^\mathrm{(t)}$ to capture surface texture patterns across objects. We enforce orthogonality between the structural features $E_t^\mathrm{(s)}$ and texture features $E_t^\mathrm{(t)}$. Specifically, we build an orthogonal loss $\mathcal{L}\mathrm{o}$ using the vectorized $F_t^\mathrm{(s)}\in\mathbb{R}^{C\times (H\times W)}$, $F_t^\mathrm{(t)}\in\mathbb{R}^{C\times (H\times W)}$ and their Gram matrices \cite{DBLP:journals/corr/GatysEB15a}, 
\begin{equation}
    \mathcal{L}_\mathrm{o}=\mathrm{cossim}(F_t^\mathrm{(s)},F_t^\mathrm{(t)})    +\mathrm{cossim}\left(\mathrm{gram}(F_t^\mathrm{(s)},\mathrm{gram}(F_t^\mathrm{(t)}\right),
\end{equation}
where $\mathrm{cossim}$ calculates the cosine similarity between features, and $\mathrm{gram}$ computes the Gram matrix of the features and vectorizes them. The reflection decoder derives the reflection component $\hat{R}_t$ from the sum of structural features $E_t^\mathrm{(s)}$ and texture features $E_t^\mathrm{(t)}$. This prevents the structural features $E_t^\mathrm{(s)}$ from containing unrelated local texture information. Experiments validate that relying on local patterns reduces robustness when cameras encounter corruption.

\subsection{Semantic-Structural Correlation}

Decoupling surface patterns improves robustness against interfering textures from scenes or cameras. However, semantic cues aid structural understanding to provide depth clues. Semantic objects in the scene carry significant structural information, with depth continuity for the same object as shown in Figure \ref{fig:sub-problems}. $I_t$ passes through the structure encoder to produce $F_t^{(\mathrm{s})}$, and through the semantic expert encoder to generate $F_t^{(\mathrm{e})}\in\mathbb{R}^{C\times (H^\prime\times W^\prime)}$. The structural embedding $F_t^{(\mathrm{s})}$ and semantic embedding $F_t^{(\mathrm{e})}$ of frame $t$ have a structural correlation that exhibits graph-like characteristics in the feature embeddings. We aim to model the structural correlation between the structural embedding $F_t^{(\mathrm{s})}$ and semantic embedding $F_t^{(\mathrm{e})}$ of frame $t$ as an isomorphic graph. This constructs feature mapping from semantic to structural cues, enabling efficient transfer of structure-relevant knowledge.

The semantic expert model extracts features $E_t^{(\mathrm{e})}$ from $I_t$. To align with $F_t^{(\mathrm{s})}$ from our structure encoder, we map $E_t^{(\mathrm{e})}$ to $F_t^{(\mathrm{e})}$ via bilinear interpolation, convolution, and vectorization. The correlation between $F_t^{(\mathrm{s})}$ and $F_t^{(\mathrm{e})}$ reflects cross-model feature relationships. We represent this via the inter-channel correlation matrix $A\left(F_t^{(\mathrm{s})},F_t^{(\mathrm{e})}\right)$, where
\begin{equation}
    A\left(F_t^{(\mathrm{s})},F_t^{(\mathrm{e})}\right)=\frac{F_t^{(\mathrm{s})}\cdot F_t^{(\mathrm{e})}}{\Vert F_t^{(\mathrm{s})}\Vert\cdot\Vert F_t^{(\mathrm{e})}\Vert}.
\end{equation}
We formulate $A\left(F_t^{(\mathrm{s})},F_t^{(\mathrm{e})}\right)$ as the adjacency matrix of a graph. By performing feature aggregation via GIN (Graph Isomorphism Network) projector \cite{DBLP:conf/iclr/XuHLJ19}, this effectively constructs an isomorphic graph between the feature nodes across the structure encoder and the semantic expert encoder. It enables the distillation of structural knowledge from the expert to the student model.

The expert encoder and structure encoder aggregate structural knowledge between the embedding nodes $F_t^{(\mathrm{e})}$ and $F_t^{(\mathrm{s})}$ to construct the virtual embedding nodes ${F_t^{(\mathrm{s})}}'$ and ${F_t^{(\mathrm{e})}}'$,
\begin{equation}
\begin{array}{c}
    {F_t^{(\mathrm{s})}}'=\mathrm{gin}\left(\theta^{(\mathrm{e\rightarrow s})},F_t^{(\mathrm{e})},A\left(F_t^{(\mathrm{e})},F_t^{(\mathrm{s})}\right)\right)\\{F_t^{(\mathrm{e})}}'=\mathrm{gin}\left(\theta^{(\mathrm{s\rightarrow e})},F_t^{(\mathrm{s})},A\left(F_t^{(\mathrm{s})},F_t^{(\mathrm{e})}\right)\right)
\end{array},
\end{equation}
where $\mathrm{gin}$ performs feature aggregation based on parameters $\theta$, embedding nodes $F_t$, and adjacency matrix $A(\cdot,\cdot)$. We update the feature nodes via two-dimensional convolutional layers in the aggregation function of GIN. The virtual nodes ${F_t^{(\mathrm{s})}}'$ and ${F_t^{(\mathrm{e})}}'$ progressively converge to the feature nodes $F_t^{(\mathrm{e})}$ and $F_t^{(\mathrm{s})}$ respectively. We define the loss function for this convergence as
\begin{equation}
    \mathcal{L}_\mathrm{d} = 2- \mathrm{cossim}\left({F_t^{(\mathrm{s})}}', F_t^{(\mathrm{s})}\right)
    -\mathrm{cossim}\left({F_t^{(\mathrm{e})}}', F_t^{(\mathrm{e})}\right)
\end{equation}
By optimizing the cosine distance between the virtual and feature nodes, the structure encoder obtains semantic structural knowledge from the expert encoder through bidirectional feature aggregation. 

\subsection{Total Loss}

Modeling depth-structure consistency, local texture disambiguation, and semantic-structural correlation constitutes multiple optimization objectives. We consolidate these goals into one overall loss $\mathcal{L}$ for joint optimization, 
\begin{equation}
    \mathcal{L}=\lambda_\mathrm{p}\mathcal{L}_\mathrm{p}+\lambda_\mathrm{v}\mathcal{L}_\mathrm{v}+\lambda_\mathrm{e}\mathcal{L}_\mathrm{e}+\lambda_\mathrm{r}\mathcal{L}_\mathrm{r}+\lambda_\mathrm{o}\mathcal{L}_\mathrm{o}+\lambda_\mathrm{d}\mathcal{L}_\mathrm{d},
\end{equation}
where $\lambda_\mathrm{p}$, $\lambda_\mathrm{v}$, $\lambda_\mathrm{e}$, $\lambda_\mathrm{r}$, $\lambda_\mathrm{o}$, $\lambda_\mathrm{d}$ balance the weights of different loss terms. The structure-centric self-supervised monocular depth model is trained with $\mathcal{L}$ as the overall optimization objective. We provide a comprehensive list of all the weights $\lambda$ used during the training process in the supplementary materials.

\section{Experiments}
\label{sec:exp}

To validate the efficacy of our structure-centric monocular depth estimation, we experimentally verify that: our method effectively incorporates scene structure information from semantics and illumination, enhancing the estimation accuracy of self-supervised depth estimation models in a wider range of real-world scenes; our proposed approach substantially improves depth estimation accuracy under common outdoor adverse weather, motion blur and in poor illumination conditions like night-time environments, realizing robust monocular depth estimation.

\subsection{Implementation Details}

We train our model on the clear KITTI dataset \cite{DBLP:journals/ijrr/GeigerLSU13} to avoid cheating under corrupted monocular depth estimation. No corrupted samples are seen during training. The original KITTI samples have a resolution of $1242\times375$. We scale them to $640\times192$ for faster training. We train our model for 20 epochs using the Adam optimizer on the \texttt{eigen\_zhou} split \cite{DBLP:journals/ijrr/GeigerLSU13} with static sequences removed.  We implement the model in PyTorch on a server with NVIDIA V100S GPUs and Intel Xeon Gold 6132 CPUs. We utilize the advanced SAM-Huge model \cite{kirillov2023segment} as our semantic expert model in the knowledge distillation process. Throughout the distillation process, we exclusively rely on commonly used semantic segmentation teacher models and do not introduce any prior depth knowledge. In our experiments, we present two model configurations: one built upon MonoDepth2, referred to as $\textrm{Ours}_\textrm{R}$, and another based on MonoViT, denoted as $\textrm{Ours}_\textrm{T}$ (distilled from SAM-Huge \cite{kirillov2023segment}). These configurations illustrate the broad applicability of our proposed training approach. The learning rates are $5\times10^{-5}$ for the structure and texture encoders of $\textrm{Ours}_\textrm{T}$, and $1\times10^{-4}$ for other components. We use exponential decay with a ratio of 0.9 for learning rate scheduling. 

\subsection{Depth Estimation Performance under Corruptions}

Adverse weather and motion blur are common camera degradation scenarios in outdoor driving or sports. For adverse weather, we select the snow, frost, fog, and rain conditions that interfere most with camera observations for validation. We validate robustness and out-of-distribution generalization on corrupted KITTI, DENSE \cite{dense-dataset} and nuScenes \cite{DBLP:conf/cvpr/CaesarBLVLXKPBB20} nighttime datasets using models trained on clear KITTI. We compare against a series of state-of-the-art models with innovations in training approaches, e.g. SCDepthV1 \cite{DBLP:journals/ijcv/BianZWLZSCR21}, SCDepthV3 \cite{DBLP:journals/corr/abs-2211-03660}, PackNet-SfM \cite{DBLP:conf/iclr/GuiziliniHLAG20}, DynaDepth \cite{10.1007/978-3-031-19839-7_9}, and SGDepth \cite{DBLP:conf/eccv/KlingnerTMF20}, and model architectures, e.g. MonoViT \cite{DBLP:conf/3dim/ZhaoZPTGZHTM22}, LiteMono \cite{DBLP:conf/cvpr/NingFGN23}, HRDepth \cite{DBLP:conf/aaai/LyuLWKLLCY21}. All of the above works using different training approaches adopt the model architecture configuration used in MonoDepth2. We favor lower Abs Rel, Sq Rel, RMSE, and RMSE log, while higher $\delta_1$, $\delta_2$, and $\delta_3$ indicate better performance for all metrics in this section. The presentation of representative indicators takes place in Table \ref{tab:kitti} and Table \ref{tab:dense}. For more detailed indicators, please refer to the supplementary materials. In the following comparisons, to ensure a fair comparison, we separately compare the models of the MonoDepth2 architecture (including $\textrm{Ours}_\textrm{R}$) and the models of the hybrid architecture of Convolutional Networks and Transformer (including $\textrm{Ours}_\textrm{T}$).

\begin{table}[!t]
    \centering
    \caption{Performance comparison of monocular depth estimation on the KITTI Snow, KITTI Frost, and KITTI Motion Blur test sets. To ensure fairness in comparisons and avoid introducing any corrupted samples, all the models mentioned above are trained on the KITTI Clear training dataset. Upper Section: ResNet-based Architectures. Lower Section: Hybrid Architectures. $\downarrow$: Lower is better. $\uparrow$: Higher is better.}
    \begin{tabular}{cc|cc|cc|cc|cc}
    \toprule
    \multirow{3}{*}{Method} & \multirow{3}{*}{Knowledge} & \multicolumn{2}{c|}{Snow} & \multicolumn{2}{c|}{Frost} & \multicolumn{2}{c|}{Motion Blur} & \multicolumn{2}{c}{Clear}\\
    && \makecell{Abs\\Rel}$\downarrow$ & $\delta_1\uparrow$ & \makecell{Abs\\Rel}$\downarrow$ & $\delta_1\uparrow$ & \makecell{Abs\\Rel}$\downarrow$ & $\delta_1\uparrow$ & \makecell{Abs\\Rel}$\downarrow$ & $\delta_1\uparrow$ \\
    \midrule
        MonoDepth2 & M & 0.462 & 0.330 & 0.289 & 0.534 & \underline{0.242} & \underline{0.613} & 0.117 & 0.875 \\ 
        SCDepthV1 & M & 0.435 & 0.335 & \underline{0.259} & 0.572 & 0.334 & 0.492 & 0.118 & 0.860 \\ 
        HRDepth & M & 0.520 & \textbf{0.478} & 0.492 & 0.504 & 0.635 & 0.415 & \textbf{0.106} & \textbf{0.891} \\
        SCDepthV3 & M+D & \underline{0.410} & 0.330 & 0.321 & 0.436 & 0.282 & 0.537 & 0.117 & 0.866 \\ 
        DynaDepth & M+I & 0.484 & 0.301 & 0.320 & 0.493 & 0.345 & 0.487 & \underline{0.114} & \underline{0.876} \\ 
        SGDepth & M+S & 0.473 & 0.322 & 0.265 & \underline{0.575} & 0.259 & 0.578 & 0.117 & 0.874 \\ 
        PackNet-SfM & M+S & 0.426 & 0.349 & 0.315 & 0.491 & 0.291 & 0.519 & 0.116 & 0.868 \\
        $\textbf{Ours}_\textbf{R}$ & M+S & \textbf{0.404} & \underline{0.358} & \textbf{0.236} & \textbf{0.608} & \textbf{0.238} & \textbf{0.617} & 0.118 & 0.869 \\ 
    \midrule
        LiteMono & M & 0.381 & 0.355 & 0.258 & 0.567 & 0.252 & 0.582 & 0.113 & 0.882 \\
        MonoViT & M & \underline{0.230} & \underline{0.621} & \underline{0.212} & \underline{0.663} & \underline{0.189} & \underline{0.699} & \textbf{0.100} & \textbf{0.898} \\ 
        $\textbf{Ours}_\textbf{T}$ & M+S & \textbf{0.224} & \textbf{0.627} & \textbf{0.202} & \textbf{0.678} & \textbf{0.183} & \textbf{0.710} & \underline{0.105} & \underline{0.894} \\ 
    \bottomrule
    \end{tabular}
    \label{tab:kitti}
\end{table}

\noindent\textbf{Robustness and Generalization.} Obtaining accurate depth data under corrupted conditions remains challenging. In the adverse real world, the robustness to corruptions reflects the model's out-of-distribution generalization ability. To reliably benchmark monocular depth methods, we build corrupted versions of the KITTI test set following the approaches in \cite{DBLP:journals/corr/abs-1907-07484} and \cite{DBLP:conf/cvpr/KarYAZ22}. As shown in Table \ref{tab:kitti}, under the benchmarks of Snow, Frost, and Motion Blur scenes constructed based on KITTI, both $\textrm{Ours}_\text{T}$ and $\textrm{Ours}_\text{R}$ have leading performances in their respective sections. Notably, $\textrm{Ours}_\text{R}$ achieves significant gains over MonoDepth2 under the same model architecture, demonstrating the efficacy of our training scheme for robustness under corruptions. Monocular depth estimation also faces challenges in handling out-of-distribution images due to camera specifics. We conduct benchmarks on the DENSE dataset and the nuScenes  nighttime dataset, which have camera configurations completely different from the KITTI dataset. These tests cover scenarios with rain, fog, clear weather, and poor nighttime lighting conditions. As shown in Table \ref{tab:dense}, under the rain, fog, and clear scenarios, both $\textrm{Ours}_\text{R}$ and $\textrm{Ours}_\text{T}$ significantly surpass other methods. This demonstrates the effectiveness of our approach in improving out-of-distribution monocular depth estimation performance in real adverse weather scenarios. The DENSE dataset also covers daytime and nighttime testing scenarios, each including testing samples of rain, fog, and clear scenarios during both daytime and nighttime. To highlight the validation in nighttime environments with poor lighting conditions and uneven light distribution, we compared benchmarks of the nighttime nuScenes dataset in Table \ref{tab:nuscenes}. Here, $\textrm{Ours}_\text{R}$ has a significant advantage over other works, and $\textrm{Ours}_\text{T}$ also has certain advantages for hybrid architecture tasks.

\begin{table}[!t]
    \centering
    \renewcommand\arraystretch{0.8}
    \tabcolsep=1.2pt
    \caption{Generalization performance comparison of monocular depth estimation on the DENSE benchmark. The benchmark comprises real rain and fog scenarios under both daytime and night conditions. Upper Section: ResNet-based Architectures. Lower Section: Hybrid Architectures. $\downarrow$: Lower is better. $\uparrow$: Higher is better.}
    \begin{tabular}{cc|ccc|ccc|ccc}
        \toprule
        \multirow{3}{*}{Method} & \multirow{3}{*}{\makecell{Know\\-ledge}} & \multicolumn{3}{c|}{Fog} & \multicolumn{3}{c|}{Rain} & \multicolumn{3}{c}{Clear} \\
        && \makecell{Abs\\Rel}$\downarrow$ & RMSE$\downarrow$ & $\delta_1\uparrow$ & \makecell{Abs\\Rel}$\downarrow$ & RMSE$\downarrow$ &  $\delta_1\uparrow$ & \makecell{Abs\\Rel}$\downarrow$ & RMSE$\downarrow$ &  $\delta_1\uparrow$ \\
        \midrule
        MonoDepth2 & M & 0.266 & 4.865 & 0.526 & 0.274 & 4.962 & 0.517 & 0.281 & 4.995 & \underline{0.572} \\ 
        SCDepthV1 & M & 0.269 & 5.125 & 0.539 & 0.266 & 5.118 & 0.537 & 0.297 & 5.147 & 0.507 \\ 
        HRDepth & M & 0.262 & 4.707 & 0.533 & 0.262 & \underline{4.730} & 0.528 & 0.272 & 4.681 & 0.532 \\ 
        SCDepthV3 & M+D & 0.257 & 5.087 & \underline{0.578} & 0.253 & 5.087 & \underline{0.581} & 0.289 & 5.152 & 0.531 \\ 
        DynaDepth & M+I & 0.271 & 5.191 & 0.519 & 0.281 & 5.311 & 0.504 & 0.273 & 4.912 & 0.493 \\ 
        SGDepth & M+S & 0.260 & 4.967 & 0.548 & \underline{0.252} & 4.912 & 0.563 & 0.261 & 4.856 & 0.547 \\ 
        PackNet-SfM & M+S & \underline{0.254} & \underline{4.697} & 0.532 & 0.256 & 4.765 & 0.533 & \underline{0.258} & \textbf{4.585} & 0.550 \\ 
        $\textbf{Ours}_\textbf{R}$ & M+S & \textbf{0.240} & \textbf{4.567} & \textbf{0.580} & \textbf{0.247} & \textbf{4.642} & \textbf{0.587} & \textbf{0.256} & \underline{4.645} & \textbf{0.575} \\
        \midrule
        LiteMono & M & 0.325 & 4.863 & 0.422 & 0.322 & 4.787 & 0.414 & 0.310 & 5.095 & 0.481 \\ 
        MonoViT & M & \underline{0.239} & \underline{4.600} & \underline{0.556} & \underline{0.237} & \underline{4.767} & \underline{0.564} & \underline{0.242} & \underline{4.320} & \underline{0.555} \\ 
        $\textbf{Ours}_\textbf{T}$ & M+S & \textbf{0.224} & \textbf{3.907} & \textbf{0.583} & \textbf{0.224} & \textbf{4.046} & \textbf{0.571} & \textbf{0.218} & \textbf{3.725} & \textbf{0.606} \\ 
        \bottomrule
    \end{tabular}
    \label{tab:dense}
\end{table}

\begin{table}[]
    \centering
    \tabcolsep=1.2pt
    \renewcommand\arraystretch{0.8}
    \caption{Generalization performance comparison of monocular depth estimation on the nuScenes nighttime benchmark. This benchmark has poor lighting conditions and is subject to a large amount of uneven lighting interference. Upper Section: ResNet-based Architectures. Lower Section: Hybrid Architectures. $\downarrow$: Lower is better. $\uparrow$: Higher is better.}
    \begin{tabular}{cc|cccc|ccc}
    \toprule
    \multirow{2}{*}{Method} & \multirow{2}{*}{Knowledge} & \makecell{Abs\\Rel}$\downarrow$ & \makecell{Sq\\Rel}$\downarrow$ & \makecell{RMSE}$\downarrow$ & \makecell{RMSE\\Log}$\downarrow$ & $\delta_1\uparrow$ & $\delta_2\uparrow$ & $\delta_3\uparrow$ \\

    \midrule
        MonoDepth2 & M & \underline{0.386}  &   \underline{5.030}  &  \underline{13.090}  &   \underline{0.548}  &   \underline{0.386}  &   \underline{0.671}  &   \textbf{0.820}  \\
        SCDepthV1 & M &   0.468  &   6.760  &  14.773  &   0.662  &   0.322  &   0.570  &   0.733  \\
        HRDepth & M &   0.458  &   6.131  &  14.439  &   0.632  &   0.305  &   0.556  &   0.739  \\
        SCDepthV3 & M+D &   0.414  &   5.560  &  14.488  &   0.627  &   0.321  &   0.613  &   0.774  \\
        DynaDepth & M+I &   0.483  &   6.820  &  14.724  &   0.643  &   0.293  &   0.559  &   0.744  \\
        SGDepth & M+S &   0.414  &   5.585  &  13.888  &   0.591  &   0.350  &   0.621  &   0.785  \\
        PackNet-SfM & M+S &   0.442  &   6.095  &  13.867  &   0.588  &   0.319  &   0.609  &   0.791  \\
        $\textbf{Ours}_\textbf{R}$ & M+S &   \textbf{0.364}  &   \textbf{4.831}  &  \textbf{12.998}  &   \textbf{0.535}  &   \textbf{0.412}  &   \textbf{0.686}  &   \underline{0.811}  \\
        \midrule
        LiteMono & M &   0.282  &   3.794  &  12.074  &   0.457  &   0.531  &   0.777  &   0.861  \\
        MonoViT & M &   \underline{0.272}  &   \textbf{3.268}  &  \textbf{10.775}  &   \textbf{0.411}  &   \underline{0.557}  &   \textbf{0.783}  &   \textbf{0.882}  \\
        $\textbf{Ours}_\textbf{T}$ & M+S &   \textbf{0.269}  &   \underline{3.416}  &  \underline{11.282}  &   \underline{0.420}  &   \textbf{0.569}  &   \underline{0.781}  &   \underline{0.872}  \\
        \bottomrule
    \end{tabular}
    \label{tab:nuscenes}
\end{table}

\noindent\textbf{Ablation Studies.} As shown in Table \ref{tab:ablation}, we conduct ablation studies on the generalizability of the training scheme, as well as the effects of local texture disambiguation and semantic-structural correlation on monocular depth robustness. Both $\textrm{Ours}_\text{R}$ and $\textrm{Ours}_\text{T}$ outperform their respective baselines, demonstrating the general applicability of our framework. Removing local texture disambiguation and semantic-structural correlation from $\textrm{Ours}_\text{T}$ reduces performance, but it still surpasses MonoViT. Removing the GIN Projector during semantic distillation led to further performance degradation, validating its role in constructing isomorphic mappings between feature channels. These results highlight the positive impact of addressing these sub-problems and the versatility of our approach in supporting semantic knowledge distillation with different ViT expert models.

\begin{table}[!t]
    \centering
    \tabcolsep=1.2pt
    \renewcommand\arraystretch{0.8}
    \caption{Ablation studies of monocular depth estimation on the KITTI Snow, KITTI Frost, KITTI Motion Blur, and KITTI Clear test sets. TI: Local texture disambiguation. SD: Semantic distillation. GP: GIN Projector. R: The architecture of MonoDepth2 based on ResNet-18 \cite{DBLP:conf/cvpr/HeZRS16}. T: The architecture of MonoViT based on MPViT-Tiny \cite{DBLP:conf/cvpr/LeeKWH22}.}
    \small
    \begin{tabular}{cccc|cc|cc|cc|cc}
        \toprule
        \multicolumn{4}{c|}{Ablation} & \multicolumn{2}{c|}{Snow} & \multicolumn{2}{c|}{Frost} & \multicolumn{2}{c|}{Motion Blur} & \multicolumn{2}{c}{Clear}\\
        \multirow{2}{*}{Arch} & \multirow{2}{*}{TI} & \multirow{2}{*}{SD} & \multirow{2}{*}{GP} & \makecell{Abs\\Rel}$\downarrow$ & $\delta_1\uparrow$ & \makecell{Abs\\Rel}$\downarrow$ & $\delta_1\uparrow$ & \makecell{Abs\\Rel}$\downarrow$ & $\delta_1\uparrow$ & \makecell{Abs\\Rel}$\downarrow$ & $\delta_1\uparrow$ \\
        \midrule
        R & \XSolid & \XSolid & \XSolid & 0.462 & 0.330 & 0.289 & 0.534 & 0.242 & 0.613 & 0.117 & 0.875 \\
        R & \Checkmark & \Checkmark & \Checkmark & 0.404 & 0.358 & 0.236 & 0.608 & 0.238 & 0.617 & 0.118 & 0.869 \\
        \midrule
        T & \XSolid & \XSolid & \XSolid & 0.230 & \underline{0.621} & 0.212 & 0.663 & 0.189 & 0.699 & \textbf{0.100} & \textbf{0.898} \\
        T & \XSolid & \Checkmark & \Checkmark & \underline{0.227} & 0.617 & 0.209 & \textbf{0.681} & 0.186 & 0.704 & \underline{0.103} & 0.893 \\
        T & \XSolid & \Checkmark & \XSolid & 0.228 & 0.590 & 0.205 & \underline{0.678} & 0.188 & 0.677 & 0.107 & 0.889 \\
        T & \Checkmark & \XSolid & \XSolid & 0.232 & 0.623 & \underline{0.204} & 0.677 & 0.185 & 0.692 & 0.109 & 0.893 \\
        T & \Checkmark & \Checkmark & \Checkmark & \textbf{0.224} & \textbf{0.627} & \textbf{0.203} & \underline{0.678} & \textbf{0.183} & \textbf{0.710} & 0.105 & \underline{0.894} \\ 
        \bottomrule
    \end{tabular}
    \label{tab:ablation}
\end{table}

\subsection{Feature Map Visualization}

\begin{figure*}[t!]
    \centering
    \includegraphics[width=0.8\textwidth]{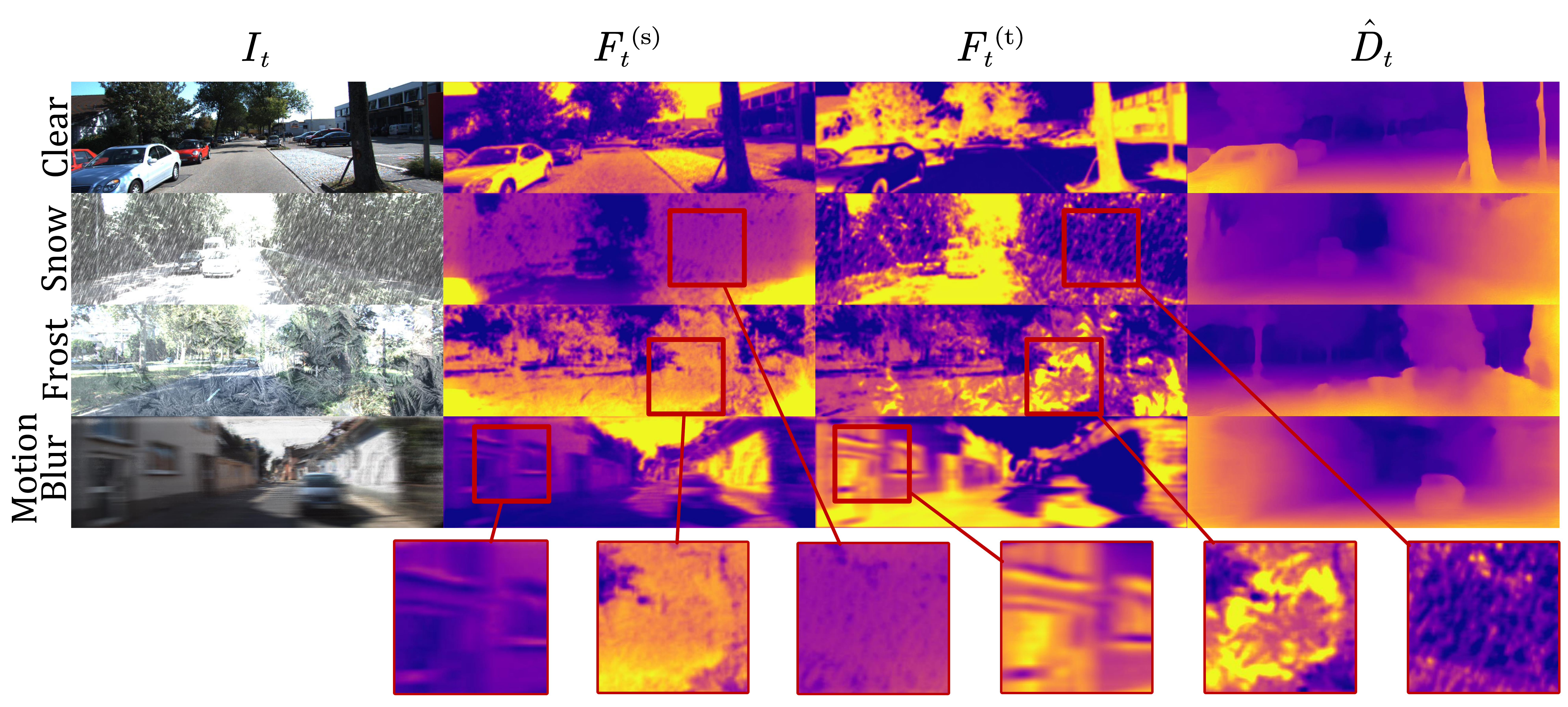}
    \caption{Visualization of intermediate feature maps. For features $F_t^{(\mathrm{s})}$ and $F_t^{(\mathrm{t})}$, yellow indicates high values. For predicted depth $\hat{D}_t$, purple denotes high values. We provide a detailed description of the computation method for visual feature maps in the supplementary materials.}
    \label{fig:visulization}
\end{figure*}

To enable the structure encoder to focus on scene structure and avoid over-reliance on local textures that degrades depth estimation under interference, we aim to solve local texture disambiguation and semantic-structural correlation. By introducing a texture encoder to model fine textures in the scene, we expect mutual suppression between the two encoders to improve the structure encoder's robustness to interfering textures. To validate this, we visualize the intermediate features $F_t^{(\mathrm{s})}$ and $F_t^{(\mathrm{t})}$ of the structure encoder and texture decoder. We upsample the multi-scale features of the encoder and merge the channels. Additionally, we perform PCA (Principal Component Analysis) dimensionality reduction on the channel dimensions to form visualizations of $F_t^{(\mathrm{s})}$ and $F_t^{(\mathrm{t})}$. As seen in Figure \ref{fig:visulization}, the reflection encoder successfully extracts interfering textures, which manifests in $F_t^{(\mathrm{t})}$ as clearer snow tracks and frost textures, and even accurately extracts radial blurring caused by motion. Given that the texture feature $F_t^{(\mathrm{t})}$ is decoupled from the structural feature $F_t^{(\mathrm{s})}$, $F_t^{(\mathrm{s})}$ clearly filters out a large amount of interfering textures. This corroborates the effectiveness of our method.

\section{Conclusion}
\label{sec:con}

We start from the principles of imaging in nature and the perspective of object textures, exploring three potential sub-problems in monocular depth estimation: local texture disambiguation and semantic-structural correlation. Current methods do not adequately model these factors, leading to significant degradation in challenging scenarios like adverse weather, motion blur, and nighttime conditions.

We approach these sub-problems by modeling them as optimizable self-supervised objectives, greatly enhancing the robustness of self-supervised monocular depth estimation under interference through knowledge distillation. We disentangle image features into scene structure and local texture, promoting a mutual suppression between the structure encoder and texture encoder to minimize over-reliance on local textures. Furthermore, learnable isomorphic graphs, coupled with a semantic expert model, facilitate the integration of structural knowledge into the depth model.

Our method achieves state-of-the-art performance under adverse weather, poor nighttime lighting, and motion blur scenarios on corrupted KITTI, DENSE, and nighttime nuScenes datasets, demonstrating its effectiveness. Visualizations show successful separation of scene structure from interfering textures, highlighting the robustness of our self-supervised approach for monocular depth estimation. We establish that not all texture details are beneficial for depth estimation; starting from global object semantics with effective guidance enhances estimation performance.

\noindent \textbf{Acknowledgments.} This work was supported in part by the Strategic Priority Research Program of Chinese Academy of Sciences under Grant XDA28040500, the National Natural Science Foundation of China under Grant 62261042, the Key Research Projects of the Joint Research Fund for Beijing Natural Science Foundation and the Fengtai Rail Transit Frontier Research Joint Fund under Grant L221003, and the Beijing Natural Science Foundation under Grant 4232035 and 4222034.


%
%
\bibliographystyle{splncs04}
\bibliography{main}

\begin{thebibliography}{10}
\providecommand{\url}[1]{\texttt{#1}}
\providecommand{\urlprefix}{URL }
\providecommand{\doi}[1]{https://doi.org/#1}

\bibitem{DBLP:journals/ijcv/BianZWLZSCR21}
Bian, J., Zhan, H., Wang, N., Li, Z., Zhang, L., Shen, C., Cheng, M., Reid, I.: Unsupervised scale-consistent depth learning from video. Int. J. Comput. Vis.  \textbf{129}(9),  2548--2564 (2021). \doi{10.1007/s11263-021-01484-6}, \url{https://doi.org/10.1007/s11263-021-01484-6}

\bibitem{DBLP:conf/nips/BianLWZSC019}
Bian, J., Li, Z., Wang, N., Zhan, H., Shen, C., Cheng, M., Reid, I.D.: Unsupervised scale-consistent depth and ego-motion learning from monocular video. In: Wallach, H.M., Larochelle, H., Beygelzimer, A., d'Alch{\'{e}}{-}Buc, F., Fox, E.B., Garnett, R. (eds.) Advances in Neural Information Processing Systems 32: Annual Conference on Neural Information Processing Systems 2019, NeurIPS 2019, December 8-14, 2019, Vancouver, BC, Canada. pp. 35--45 (2019), \url{https://proceedings.neurips.cc/paper/2019/hash/6364d3f0f495b6ab9dcf8d3b5c6e0b01-Abstract.html}

\bibitem{DBLP:conf/cvpr/CaesarBLVLXKPBB20}
Caesar, H., Bankiti, V., Lang, A.H., Vora, S., Liong, V.E., Xu, Q., Krishnan, A., Pan, Y., Baldan, G., Beijbom, O.: nuscenes: {A} multimodal dataset for autonomous driving. In: 2020 {IEEE/CVF} Conference on Computer Vision and Pattern Recognition, {CVPR} 2020, Seattle, WA, USA, June 13-19, 2020. pp. 11618--11628. Computer Vision Foundation / {IEEE} (2020). \doi{10.1109/CVPR42600.2020.01164}, \url{https://openaccess.thecvf.com/content\_CVPR\_2020/html/Caesar\_nuScenes\_A\_Multimodal\_Dataset\_for\_Autonomous\_Driving\_CVPR\_2020\_paper.html}

\bibitem{Chen2023}
Chen, Y., Wen, C., Liu, W., He, W.: A depth iterative illumination estimation network for low-light image enhancement based on retinex theory. Scientific Reports  \textbf{13}(1),  19709 (Nov 2023). \doi{10.1038/s41598-023-46693-w}, \url{https://doi.org/10.1038/s41598-023-46693-w}

\bibitem{10306272}
Gao, W., Rao, D., Yang, Y., Chen, J.: Edge devices friendly self-supervised monocular depth estimation via knowledge distillation. IEEE Robotics and Automation Letters  \textbf{8}(12),  8470--8477 (2023). \doi{10.1109/LRA.2023.3330054}

\bibitem{DBLP:conf/iccv/GasperiniMJNT23}
Gasperini, S., Morbitzer, N., Jung, H., Navab, N., Tombari, F.: Robust monocular depth estimation under challenging conditions. In: {IEEE/CVF} International Conference on Computer Vision, {ICCV} 2023, Paris, France, October 1-6, 2023. pp. 8143--8152. {IEEE} (2023). \doi{10.1109/ICCV51070.2023.00751}, \url{https://doi.org/10.1109/ICCV51070.2023.00751}

\bibitem{DBLP:journals/corr/GatysEB15a}
Gatys, L.A., Ecker, A.S., Bethge, M.: A neural algorithm of artistic style. CoRR  \textbf{abs/1508.06576} (2015), \url{http://arxiv.org/abs/1508.06576}

\bibitem{DBLP:journals/ijrr/GeigerLSU13}
Geiger, A., Lenz, P., Stiller, C., Urtasun, R.: Vision meets robotics: The {KITTI} dataset. Int. J. Robotics Res.  \textbf{32}(11),  1231--1237 (2013). \doi{10.1177/0278364913491297}, \url{https://doi.org/10.1177/0278364913491297}

\bibitem{DBLP:conf/cvpr/GodardAB17}
Godard, C., Aodha, O.M., Brostow, G.J.: Unsupervised monocular depth estimation with left-right consistency. In: 2017 {IEEE} Conference on Computer Vision and Pattern Recognition, {CVPR} 2017, Honolulu, HI, USA, July 21-26, 2017. pp. 6602--6611. {IEEE} Computer Society (2017). \doi{10.1109/CVPR.2017.699}, \url{https://doi.org/10.1109/CVPR.2017.699}

\bibitem{DBLP:conf/iccv/GodardAFB19}
Godard, C., Aodha, O.M., Firman, M., Brostow, G.J.: Digging into self-supervised monocular depth estimation. In: 2019 {IEEE/CVF} International Conference on Computer Vision, {ICCV} 2019, Seoul, Korea (South), October 27 - November 2, 2019. pp. 3827--3837. {IEEE} (2019). \doi{10.1109/ICCV.2019.00393}, \url{https://doi.org/10.1109/ICCV.2019.00393}

\bibitem{GOYAL2024102151}
Goyal, B., Dogra, A., Lepcha, D.C., Goyal, V., Alkhayyat, A., Chohan, J.S., Kukreja, V.: Recent advances in image dehazing: Formal analysis to automated approaches. Information Fusion  \textbf{104},  102151 (2024). \doi{https://doi.org/10.1016/j.inffus.2023.102151}, \url{https://www.sciencedirect.com/science/article/pii/S1566253523004670}

\bibitem{dense-dataset}
Gruber, T., Bijelic, M., Heide, F., Ritter, W., Dietmayer, K.: Pixel-accurate depth evaluation in realistic driving scenarios. In: 2019 International Conference on 3D Vision (3DV). pp. 95--105 (2019). \doi{10.1109/3DV.2019.00020}

\bibitem{DBLP:conf/iclr/GuiziliniHLAG20}
Guizilini, V., Hou, R., Li, J., Ambrus, R., Gaidon, A.: Semantically-guided representation learning for self-supervised monocular depth. In: 8th International Conference on Learning Representations, {ICLR} 2020, Addis Ababa, Ethiopia, April 26-30, 2020. OpenReview.net (2020), \url{https://openreview.net/forum?id=ByxT7TNFvH}

\bibitem{DBLP:conf/cvpr/HeZRS16}
He, K., Zhang, X., Ren, S., Sun, J.: Deep residual learning for image recognition. In: 2016 {IEEE} Conference on Computer Vision and Pattern Recognition, {CVPR} 2016, Las Vegas, NV, USA, June 27-30, 2016. pp. 770--778. {IEEE} Computer Society (2016). \doi{10.1109/CVPR.2016.90}, \url{https://doi.org/10.1109/CVPR.2016.90}

\bibitem{DBLP:conf/iclr/HendrycksD19}
Hendrycks, D., Dietterich, T.G.: Benchmarking neural network robustness to common corruptions and perturbations. In: 7th International Conference on Learning Representations, {ICLR} 2019, New Orleans, LA, USA, May 6-9, 2019. OpenReview.net (2019), \url{https://openreview.net/forum?id=HJz6tiCqYm}

\bibitem{9578162}
Hoyer, L., Dai, D., Chen, Y., Köring, A., Saha, S., Van~Gool, L.: Three ways to improve semantic segmentation with self-supervised depth estimation. In: 2021 IEEE/CVF Conference on Computer Vision and Pattern Recognition (CVPR). pp. 11125--11135 (2021). \doi{10.1109/CVPR46437.2021.01098}

\bibitem{DBLP:conf/cvpr/KarYAZ22}
Kar, O.F., Yeo, T., Atanov, A., Zamir, A.: 3d common corruptions and data augmentation. In: {IEEE/CVF} Conference on Computer Vision and Pattern Recognition, {CVPR} 2022, New Orleans, LA, USA, June 18-24, 2022. pp. 18941--18952. {IEEE} (2022). \doi{10.1109/CVPR52688.2022.01839}, \url{https://doi.org/10.1109/CVPR52688.2022.01839}

\bibitem{kirillov2023segment}
Kirillov, A., Mintun, E., Ravi, N., Mao, H., Rolland, C., Gustafson, L., Xiao, T., Whitehead, S., Berg, A.C., Lo, W., Doll{\'{a}}r, P., Girshick, R.B.: Segment anything. In: {IEEE/CVF} International Conference on Computer Vision, {ICCV} 2023, Paris, France, October 1-6, 2023. pp. 3992--4003. {IEEE} (2023). \doi{10.1109/ICCV51070.2023.00371}, \url{https://doi.org/10.1109/ICCV51070.2023.00371}

\bibitem{DBLP:conf/eccv/KlingnerTMF20}
Klingner, M., Term{\"{o}}hlen, J., Mikolajczyk, J., Fingscheidt, T.: Self-supervised monocular depth estimation: Solving the dynamic object problem by semantic guidance. In: Vedaldi, A., Bischof, H., Brox, T., Frahm, J. (eds.) Computer Vision - {ECCV} 2020 - 16th European Conference, Glasgow, UK, August 23-28, 2020, Proceedings, Part {XX}. Lecture Notes in Computer Science, vol. 12365, pp. 582--600. Springer (2020). \doi{10.1007/978-3-030-58565-5\_35}, \url{https://doi.org/10.1007/978-3-030-58565-5\_35}

\bibitem{DBLP:conf/nips/KongXHNCO23}
Kong, L., Xie, S., Hu, H., Ng, L.X., Cottereau, B., Ooi, W.T.: Robodepth: Robust out-of-distribution depth estimation under corruptions. In: Oh, A., Naumann, T., Globerson, A., Saenko, K., Hardt, M., Levine, S. (eds.) Advances in Neural Information Processing Systems 36: Annual Conference on Neural Information Processing Systems 2023, NeurIPS 2023, New Orleans, LA, USA, December 10 - 16, 2023 (2023), \url{http://papers.nips.cc/paper\_files/paper/2023/hash/43119db5d59f07cc08fca7ba6820179a-Abstract-Datasets\_and\_Benchmarks.html}

\bibitem{DBLP:conf/cvpr/LeeKWH22}
Lee, Y., Kim, J., Willette, J., Hwang, S.J.: Mpvit: Multi-path vision transformer for dense prediction. In: {IEEE/CVF} Conference on Computer Vision and Pattern Recognition, {CVPR} 2022, New Orleans, LA, USA, June 18-24, 2022. pp. 7277--7286. {IEEE} (2022). \doi{10.1109/CVPR52688.2022.00714}, \url{https://doi.org/10.1109/CVPR52688.2022.00714}

\bibitem{Lopez-Rodriguez2023}
Lopez-Rodriguez, A., Mikolajczyk, K.: Desc: Domain adaptation for depth estimation via semantic consistency. International Journal of Computer Vision  \textbf{131}(3),  752--771 (Mar 2023). \doi{10.1007/s11263-022-01718-1}, \url{https://doi.org/10.1007/s11263-022-01718-1}

\bibitem{10322131}
Luginov, A., Makarov, I.: Swiftdepth: An efficient hybrid cnn-transformer model for self-supervised monocular depth estimation on mobile devices. In: 2023 IEEE International Symposium on Mixed and Augmented Reality Adjunct (ISMAR-Adjunct). pp. 642--647 (2023). \doi{10.1109/ISMAR-Adjunct60411.2023.00137}

\bibitem{DBLP:conf/aaai/LyuLWKLLCY21}
Lyu, X., Liu, L., Wang, M., Kong, X., Liu, L., Liu, Y., Chen, X., Yuan, Y.: Hr-depth: High resolution self-supervised monocular depth estimation. In: Thirty-Fifth {AAAI} Conference on Artificial Intelligence, {AAAI} 2021, Thirty-Third Conference on Innovative Applications of Artificial Intelligence, {IAAI} 2021, The Eleventh Symposium on Educational Advances in Artificial Intelligence, {EAAI} 2021, Virtual Event, February 2-9, 2021. pp. 2294--2301. {AAAI} Press (2021), \url{https://ojs.aaai.org/index.php/AAAI/article/view/16329}

\bibitem{DBLP:journals/corr/abs-1907-07484}
Michaelis, C., Mitzkus, B., Geirhos, R., Rusak, E., Bringmann, O., Ecker, A.S., Bethge, M., Brendel, W.: Benchmarking robustness in object detection: Autonomous driving when winter is coming. CoRR  \textbf{abs/1907.07484} (2019), \url{http://arxiv.org/abs/1907.07484}

\bibitem{MING202114}
Ming, Y., Meng, X., Fan, C., Yu, H.: Deep learning for monocular depth estimation: A review. Neurocomputing  \textbf{438},  14--33 (2021). \doi{https://doi.org/10.1016/j.neucom.2020.12.089}, \url{https://www.sciencedirect.com/science/article/pii/S0925231220320014}

\bibitem{DBLP:conf/iccv/RanftlBK21}
Ranftl, R., Bochkovskiy, A., Koltun, V.: Vision transformers for dense prediction. In: 2021 {IEEE/CVF} International Conference on Computer Vision, {ICCV} 2021, Montreal, QC, Canada, October 10-17, 2021. pp. 12159--12168. {IEEE} (2021). \doi{10.1109/ICCV48922.2021.01196}, \url{https://doi.org/10.1109/ICCV48922.2021.01196}

\bibitem{DBLP:conf/cvpr/RematasF20}
Rematas, K., Ferrari, V.: Neural voxel renderer: Learning an accurate and controllable rendering tool. In: 2020 {IEEE/CVF} Conference on Computer Vision and Pattern Recognition, {CVPR} 2020, Seattle, WA, USA, June 13-19, 2020. pp. 5416--5426. Computer Vision Foundation / {IEEE} (2020). \doi{10.1109/CVPR42600.2020.00546}, \url{https://openaccess.thecvf.com/content\_CVPR\_2020/html/Rematas\_Neural\_Voxel\_Renderer\_Learning\_an\_Accurate\_and\_Controllable\_Rendering\_Tool\_CVPR\_2020\_paper.html}

\bibitem{DBLP:conf/iccv/SaundersVM23}
Saunders, K., Vogiatzis, G., Manso, L.J.: Self-supervised monocular depth estimation: Let's talk about the weather. In: {IEEE/CVF} International Conference on Computer Vision, {ICCV} 2023, Paris, France, October 1-6, 2023. pp. 8873--8883. {IEEE} (2023). \doi{10.1109/ICCV51070.2023.00818}, \url{https://doi.org/10.1109/ICCV51070.2023.00818}

\bibitem{10377594}
Shi, X., Dikov, G., Reitmayr, G., Kim, T.K., Ghafoorian, M.: 3d distillation: Improving self-supervised monocular depth estimation on reflective surfaces. In: 2023 IEEE/CVF International Conference on Computer Vision (ICCV). pp. 9099--9109 (2023). \doi{10.1109/ICCV51070.2023.00838}

\bibitem{DBLP:journals/corr/abs-2211-03660}
Sun, L., Bian, J., Zhan, H., Yin, W., Reid, I.D., Shen, C.: Sc-depthv3: Robust self-supervised monocular depth estimation for dynamic scenes. {IEEE} Trans. Pattern Anal. Mach. Intell.  \textbf{46}(1),  497--508 (2024). \doi{10.1109/TPAMI.2023.3322549}, \url{https://doi.org/10.1109/TPAMI.2023.3322549}

\bibitem{DBLP:conf/iccv/WangZY0X0021}
Wang, K., Zhang, Z., Yan, Z., Li, X., Xu, B., Li, J., Yang, J.: Regularizing nighttime weirdness: Efficient self-supervised monocular depth estimation in the dark. In: 2021 {IEEE/CVF} International Conference on Computer Vision, {ICCV} 2021, Montreal, QC, Canada, October 10-17, 2021. pp. 16035--16044. {IEEE} (2021). \doi{10.1109/ICCV48922.2021.01575}, \url{https://doi.org/10.1109/ICCV48922.2021.01575}

\bibitem{DBLP:conf/iclr/XuHLJ19}
Xu, K., Hu, W., Leskovec, J., Jegelka, S.: How powerful are graph neural networks? In: 7th International Conference on Learning Representations, {ICLR} 2019, New Orleans, LA, USA, May 6-9, 2019. OpenReview.net (2019), \url{https://openreview.net/forum?id=ryGs6iA5Km}

\bibitem{DBLP:conf/cvpr/NingFGN23}
Zhang, N., Nex, F., Vosselman, G., Kerle, N.: Lite-mono: A lightweight cnn and transformer architecture for self-supervised monocular depth estimation. In: {IEEE/CVF} Conference on Computer Vision and Pattern Recognition, {CVPR} 2023, Vancouver, Canada, June 18-22, 2023. {IEEE} (2023)

\bibitem{10.1007/978-3-031-19839-7_9}
Zhang, S., Zhang, J., Tao, D.: Towards scale-aware, robust, and generalizable unsupervised monocular depth estimation by integrating imu motion dynamics. In: Computer Vision – ECCV 2022: 17th European Conference, Tel Aviv, Israel, October 23–27, 2022, Proceedings, Part XXXVIII. p. 143–160. Springer-Verlag, Berlin, Heidelberg (2022). \doi{10.1007/978-3-031-19839-7_9}, \url{https://doi.org/10.1007/978-3-031-19839-7_9}

\bibitem{DBLP:conf/3dim/ZhaoZPTGZHTM22}
Zhao, C., Zhang, Y., Poggi, M., Tosi, F., Guo, X., Zhu, Z., Huang, G., Tang, Y., Mattoccia, S.: Monovit: Self-supervised monocular depth estimation with a vision transformer. In: International Conference on 3D Vision, 3DV 2022, Prague, Czech Republic, September 12-16, 2022. pp. 668--678. {IEEE} (2022). \doi{10.1109/3DV57658.2022.00077}, \url{https://doi.org/10.1109/3DV57658.2022.00077}

\bibitem{DBLP:journals/tci/ZhaoGFK17}
Zhao, H., Gallo, O., Frosio, I., Kautz, J.: Loss functions for image restoration with neural networks. {IEEE} Trans. Computational Imaging  \textbf{3}(1),  47--57 (2017). \doi{10.1109/TCI.2016.2644865}, \url{https://doi.org/10.1109/TCI.2016.2644865}

\bibitem{DBLP:conf/cvpr/ZhouBSL17}
Zhou, T., Brown, M., Snavely, N., Lowe, D.G.: Unsupervised learning of depth and ego-motion from video. In: 2017 {IEEE} Conference on Computer Vision and Pattern Recognition, {CVPR} 2017, Honolulu, HI, USA, July 21-26, 2017. pp. 6612--6619. {IEEE} Computer Society (2017). \doi{10.1109/CVPR.2017.700}, \url{https://doi.org/10.1109/CVPR.2017.700}

\end{thebibliography}

\end{document}


\title{Structure-Centric Robust Monocular Depth Estimation via Knowledge Distillation:\\Supplementary Material} 

\titlerunning{Scent-Depth: Supplementary Material}

\author{}

\authorrunning{}

\institute{}

\maketitle
\appendix


In this supplementary material, we expand upon the main text by offering clarifications on its content, presenting a comprehensive set of experimental results, and including enhanced visual representations of feature extraction and depth estimation processes. This material serves to deepen the reader's understanding and provide a more thorough insight into the methodologies and outcomes discussed in the primary research.

\section{Clarification on sub-problems}

Our work addresses three sub-problems in monocular scene structure: \textit{depth structure consistency}, \textit{local texture disambiguation}, and \textit{semantic and structural correlation}. Depth structure consistency is a fundamental problem that monocular depth estimation (MDE) needs to address. We clarify the issues of semantic and structure correlation that many previous works mentioned. Building on this foundation, we propose the depth disambiguation problem of local texture. Our main contributions are: observing the \textbf{influence of ambient lighting on scene structure} and using it to guide depth disambiguation of local texture; modeling the embedding \textbf{correlation between semantic features and scene structure features} through isomorphic graph knowledge distillation. 

\section{Trade-offs between clear and challenging performances}

Further evaluation on the public datasets DENSE and DIODE confirms that our model outperforms baseline methods, which primarily address depth consistency, such as MonoDepth2 and MonoViT, in terms of accuracy on clear test sets. The comparative results are detailed in Tables \ref{tab:diode} and \ref{tab:dense-clear}.

\begin{table}[htbp!]
    \vspace{-1em}
    \centering
    \caption{\textbf{Generalization performance comparison of MDE on the DIODE benchmark.} To ensure fairness in comparisons and avoid introducing any corrupted samples, all the models mentioned above are trained on the KITTI Clear training dataset. Upper Section: ResNet-based Architectures. Lower Section: Hybrid Architectures. $\downarrow$: Lower is better. $\uparrow$: Higher is better. (SCDepthV3 employs an additional supervised depth estimation teacher model to provide labels.)}
    \begin{tabular}{cc|cccc|ccc}
    \toprule
    \multirow{2}{*}{Method} & \multirow{2}{*}{Knowledge} & \makecell{Abs\\Rel}$\downarrow$ & \makecell{Sq\\Rel}$\downarrow$ & \makecell{RMSE}$\downarrow$ & \makecell{RMSE\\Log}$\downarrow$ & $\delta_1\uparrow$ & $\delta_2\uparrow$ & $\delta_3\uparrow$ \\
    \midrule
        MonoDepth2 & M &   0.417  &   6.041  &   6.593  &   0.427  &   0.541  &   0.769  &   0.880  \\
        SCDepthV1 & M &   0.416  &   \underline{5.807}  &   \underline{6.196}  &   0.430  &   0.518  &   0.765  &   0.886  \\
        DynaDepth & M &   0.453  &   6.671  &   6.727  &   0.446  &   0.517  &   0.756  &   0.871  \\
        HRDepth & M &   0.433  &   6.810  &   6.655  &   0.439  &   0.522  &   0.758  &   0.873  \\
        SCDepthV3 & M+D &   \textbf{0.365}  &   \textbf{4.370}  &   \textbf{5.526}  &   \textbf{0.386}  &   \textbf{0.576}  &   \textbf{0.808}  &   \textbf{0.907}  \\
        SGDepth & M+I &   0.459  &   7.409  &   7.160  &   0.459  &   0.503  &   0.741  &   0.863  \\
        PackNet-SfM & M+S &   0.443  &   6.345  &   6.664  &   0.449  &   0.507  &   0.749  &   0.868  \\
        $\mathbf{Ours}_\mathrm{R}$ & M+S &   \underline{0.414}  &   6.011  &   6.381  &   \underline{0.426}  &   \underline{0.542}  &   \underline{0.771}  &   \underline{0.882}  \\
        \midrule
        MonoViT & M &   \underline{0.348}  &   \underline{3.750}  &   \textbf{5.006}  &   \underline{0.367}  &   \underline{0.599}  &   \textbf{0.825}  &   \underline{0.911}  \\
        LiteMono & M &   0.454  &   6.598  &   6.724  &   0.465  &   0.486  &   0.734  &   0.855  \\
        $\mathbf{Ours}_\mathrm{T}$ & M+S &   \textbf{0.340}  &   \textbf{3.708}  &   \underline{5.118}  &   \textbf{0.366}  &   \underline{0.595}  &   \textbf{0.825}  &   \textbf{0.916}  \\
        \bottomrule
    \end{tabular}
    \label{tab:diode}
\end{table}

\begin{table}[!thbp]
    \centering
    \renewcommand\arraystretch{0.8}
    \caption{Detailed performance comparison of monocular depth estimation on the DENSE Clear test set. To ensure fairness in comparisons and avoid introducing any corrupted samples, all the models mentioned above are trained on the KITTI Clear training dataset. Upper Section: ResNet-based Architectures. Lower Section: Hybrid Architectures. $\downarrow$: Lower is better. $\uparrow$: Higher is better.}
    \begin{tabular}{cc|cccc|ccc}
    \toprule
    \multirow{2}{*}{Method} & \multirow{2}{*}{Knowledge} & \makecell{Abs\\Rel}$\downarrow$ & \makecell{Sq\\Rel}$\downarrow$ & \makecell{RMSE}$\downarrow$ & \makecell{RMSE\\Log}$\downarrow$ & $\delta_1\uparrow$ & $\delta_2\uparrow$ & $\delta_3\uparrow$ \\
    \midrule 
        MonoDepth2 & M &   0.281  &   1.784  &   4.995  &   0.383  &   0.572  &   0.780  &   0.889  \\
        SCDepthV1 & M &   0.297  &   1.822  &   5.147  &   0.417  &   0.507  &   0.724  &   0.869  \\
        HRDepth & M &   0.272  &   1.501  &   4.681  &   0.366  &   0.532  &   0.785  &   0.923  \\
        SCDepthV3 & M+D &   0.289  &   1.837  &   5.152  &   0.397  &   0.531  &   0.750  &   0.882  \\
        DynaDepth & M+I &   0.273  &   1.506  &   4.912  &   0.386  &   0.493  &   0.774  &   0.900  \\
        SGDepth & M+S &   0.261  &   1.459  &   4.856  &   0.368  &   0.547  &   0.794  &   0.911  \\
        PackNet-SfM & M+S &   0.258  &   1.369  &   4.585  &   0.357  &   0.550  &   0.800  &   0.917  \\
        $\textbf{Ours}_\textbf{R}$ & M+S &   0.256  &   1.445  &   4.645  &   0.354  &   0.575  &   0.815  &   0.911  \\
        \midrule
        MonoViT & M &   0.242  &   1.152  &   4.320  &   0.330  &   0.555  &   0.814  &   0.944  \\
        LiteMono & M &   0.310  &   2.096  &   5.095  &   0.367  &   0.481  &   0.802  &   0.921  \\
        $\textbf{Ours}_\textbf{T}$ & M+S &   0.218  &   0.915  &   3.725  &   0.298  &   0.606  &   0.849  &   0.961  \\
        \bottomrule
    \end{tabular}
    \label{tab:dense-clear}
\end{table}

\section{Performance Metrics Detailed Explanation}
\label{sec:metrics}

In this section, we elaborate on the computation methodologies for all metrics utilized within our research, aiming to provide a comprehensive understanding of how performance is quantified. Our evaluation framework prioritizes lower values for Absolute Relative Difference (Abs Rel), Squared Relative Difference (Sq Rel), Root Mean Squared Error (RMSE), and Root Mean Squared Error in log scale (RMSE log), while higher values of $\delta_1$, $\delta_2$, and $\delta_3$ thresholds denote superior performance.

\noindent\textbf{Absolute Relative Difference.} The Absolute Relative Difference quantifies the average relative difference between the estimated depth values and the ground truth, defined as 
\begin{equation}
    \mathcal{E}_\textrm{AbsRel}= \frac{1}{N} \sum_{i=1}^{N} \frac{|\hat{D}_t^i - D_t^i|}{D_t^i},
\end{equation}
where $D_t^i$ represents the ground truth depth for the $i^{th}$ pixel, $\hat{D}_t^i$ is the estimated depth, and $N$ denotes the total number of instances in the dataset.

\noindent\textbf{Squared Relative Difference.} The Squared Relative Difference emphasizes larger errors more significantly by computing the average of the squared relative differences between the predicted and actual depth values as
\begin{equation}
    \mathcal{E}_\textrm{SqRel} = \frac{1}{N} \sum_{i=1}^{N} \frac{(\hat{D}_t^i - D_t^i)^2}{D_t^i}.
\end{equation}

\noindent\textbf{Root Mean Squared Error.} The RMSE is a standard metric for evaluating the accuracy of predictions, calculating the square root of the average of the squared differences between estimated and true values:
\begin{equation}
    \mathcal{E}_\textrm{RMSE} = \sqrt{\frac{1}{N} \sum_{i=1}^{N} (\hat{D}_t^i - D_t^i)^2}.
\end{equation}

\noindent\textbf{Root Mean Squared Error Log.} The RMSE log applies a logarithmic transformation prior to computing the squared differences, calculated as
\begin{equation}
    \mathcal{E}_\textrm{RMSELog} = \sqrt{\frac{1}{N} \sum_{i=1}^{N} (\log(\hat{D}_t^i + 1) - \log(D_t^i + 1))^2}.
\end{equation}

\noindent\textbf{Accuracy under Thresholds.} These thresholds measure the percentage of predictions that fall within specified factor thresholds of the actual depth values, aiming to capture the accuracy of the depth estimation, defined as:
\begin{equation}
    \delta_r = \frac{1}{N} \sum_{i=1}^{N} \left( \max\left(\frac{D_t^i}{\hat{D}_t^i}, \frac{\hat{D}_t^i}{D_t^i}\right) < 1.25^r \right),
\end{equation}
where $r \in \{1,2,3\}$, corresponding to the thresholds $1.25, 1.25^2$, and $1.25^3$ respectively. A prediction is considered accurate if its ratio to the ground truth is less than the threshold value.

By leveraging these metrics, we ensure a holistic evaluation of our model's performance, capturing both the magnitude and distribution of the errors, as well as the precision of the depth estimations. 

\section{Overview of Additional Knowledge Forms}

\begin{table}[tbp]
    \centering
    \begin{minipage}{0.55\textwidth}
        \centering
        \caption{Knowledge Form Abbreviations}
        \begin{tabular}{c|c}
            \toprule
            Abbr. & Knowledge Form \\
            \midrule
            M & Monocular Camera\\
            D & Pseudo Depth Map\\
            I & Inertial Measurement Unit\\
            S & Implicit \& Explicit Semantics\\
            \bottomrule
        \end{tabular}
        \label{tab:knowledge_abbr}
    \end{minipage}\hfill
    \begin{minipage}{0.4\textwidth}
        \centering
        \caption{The weights for balancing the losses, including $\lambda_\mathrm{p}$, $\lambda_\mathrm{v}$, $\lambda_\mathrm{e}$, $\lambda_\mathrm{r}$, $\lambda_\mathrm{o}$, and $\lambda_\mathrm{d}$.}
        \begin{tabular}{c|c}
            \toprule
            $\lambda$ & Value \\
            \midrule
            $\lambda_\mathrm{p}$, $\lambda_\mathrm{e}$ & 1.0\\
            $\lambda_\mathrm{v}$, $\lambda_\mathrm{r}$ & 0.1\\
            $\lambda_\mathrm{o}$, $\lambda_\mathrm{d}$ & 0.001\\
            \bottomrule
        \end{tabular}
        \label{tab:losses}
    \end{minipage}
\end{table}

In our experimental comparison tables, we annotate the types of knowledge forms utilized by the methods, with the detailed descriptions of these knowledge forms presented in Table \ref{tab:knowledge_abbr}. Our method uniquely uses implicit semantic information, unlike other related works that depend on explicitly annotated semantic labels for training constraints.

\section{Weights of Losses}

The table in \ref{tab:losses} shows the specific values for $\lambda_\mathrm{p}$, $\lambda_\mathrm{v}$, $\lambda_\mathrm{e}$, $\lambda_\mathrm{r}$, $\lambda_\mathrm{o}$, and $\lambda_\mathrm{d}$ that we employ. Our primary objective remains to guide the model in achieving depth estimation, hence we prioritize the weights for depth structure consistency. To further guide the estimation of illumination, we set these weights higher than those for feature decoupling or distillation loss. Setting overly large weights on losses that affect features can disrupt the effective training of depth estimation.

\section{Detailed Performace Comparion}

In the main text, due to space constraints, we have selected only a subset of the most representative metrics. For the sake of completeness, we provide a detailed performance comparison of all metrics in the supplementary material.

\begin{table}[!thbp]
    \centering
    \renewcommand\arraystretch{0.8}
    \caption{Detailed performance comparison of monocular depth estimation on the KITTI Snow test set. To ensure fairness in comparisons and avoid introducing any corrupted samples, all the models mentioned above are trained on the KITTI Clear training dataset. Upper Section: ResNet-based Architectures. Lower Section: Hybrid Architectures. $\downarrow$: Lower is better. $\uparrow$: Higher is better.}
    \begin{tabular}{cc|cccc|ccc}
    \toprule
    \multirow{2}{*}{Method} & \multirow{2}{*}{Knowledge} & \makecell{Abs\\Rel}$\downarrow$ & \makecell{Sq\\Rel}$\downarrow$ & \makecell{RMSE}$\downarrow$ & \makecell{RMSE\\Log}$\downarrow$ & $\delta_1\uparrow$ & $\delta_2\uparrow$ & $\delta_3\uparrow$ \\

    \midrule 
        MonoDepth2 & M &   0.462  &   6.309  &  11.397  &   0.537  &   0.330  &   0.608  &   0.796  \\
        SCDepthV1 & M &   0.435  &   4.962  &  10.871  &   0.528  &   0.335  &   0.619  &   0.804  \\
        HRDepth & M &   0.460  &   5.826  &  10.749  &   0.538  &   0.331  &   0.605  &   0.790  \\
        SCDepthV3 & M+D &   0.410  &   4.159  &  10.721  &   0.523  &   0.330  &   0.619  &   0.816  \\
        DynaDepth & M+I &   0.484  &   6.179  &  11.449  &   0.571  &   0.301  &   0.571  &   0.768  \\
        SGDepth & M+S &   0.473  &   6.487  &  11.451  &   0.541  &   0.322  &   0.606  &   0.790  \\
        PackNet-SfM & M+S &   0.426  &   5.372  &  10.879  &   0.509  &   0.349  &   0.638  &   0.820  \\
        $\textbf{Ours}_\textbf{R}$ & M+S &   0.404  &   4.454  &  10.093  &   0.490  &   0.358  &   0.660  &   0.841  \\
        \midrule
        LiteMono & M &   0.381  &   3.858  &   9.794  &   0.476  &   0.355  &   0.675  &   0.859  \\
        MonoViT & M &   0.230  &   1.893  &   6.951  &   0.316  &   0.621  &   0.858  &   0.942  \\
        $\textbf{Ours}_\textbf{T}$ & M+S &   0.224  &   1.790  &   6.899  &   0.314  &   0.627  &   0.859  &   0.942  \\
        \bottomrule
    \end{tabular}
    \label{tab:kitti-snow}
\end{table}

\begin{table}[!thbp]
    \centering
    \renewcommand\arraystretch{0.8}
    \caption{Detailed performance comparison of monocular depth estimation on the KITTI Frost test set. To ensure fairness in comparisons and avoid introducing any corrupted samples, all the models mentioned above are trained on the KITTI Clear training dataset. Upper Section: ResNet-based Architectures. Lower Section: Hybrid Architectures. $\downarrow$: Lower is better. $\uparrow$: Higher is better.}
    \begin{tabular}{cc|cccc|ccc}
    \toprule
    \multirow{2}{*}{Method} & \multirow{2}{*}{Knowledge} & \makecell{Abs\\Rel}$\downarrow$ & \makecell{Sq\\Rel}$\downarrow$ & \makecell{RMSE}$\downarrow$ & \makecell{RMSE\\Log}$\downarrow$ & $\delta_1\uparrow$ & $\delta_2\uparrow$ & $\delta_3\uparrow$ \\
    \midrule 
        MonoDepth2 & M &   0.289  &   2.926  &   8.181  &   0.376  &   0.534  &   0.795  &   0.908  \\
        SCDepthV1 & M &   0.259  &   2.284  &   7.803  &   0.369  &   0.572  &   0.809  &   0.911  \\
        HRDepth & M &   0.353  &   3.719  &   9.007  &   0.455  &   0.452  &   0.703  &   0.853  \\
        SCDepthV3 & M+D &   0.321  &   2.897  &   9.223  &   0.432  &   0.436  &   0.741  &   0.887  \\
        DynaDepth & M+I &   0.320  &   3.307  &   8.600  &   0.424  &   0.493  &   0.747  &   0.878  \\
        SGDepth & M+S &   0.265  &   2.516  &   7.573  &   0.349  &   0.575  &   0.821  &   0.923  \\
        PackNet-SfM & M+S &   0.315  &   3.357  &   8.704  &   0.398  &   0.491  &   0.767  &   0.894  \\
        $\textbf{Ours}_\textbf{R}$ & M+S &   0.236  &   1.994  &   7.266  &   0.326  &   0.608  &   0.855  &   0.939  \\
        \midrule
        LiteMono & M &   0.258  &   2.152  &   7.336  &   0.331  &   0.567  &   0.835  &   0.940  \\
        MonoViT & M &   0.212  &   1.759  &   6.717  &   0.289  &   0.663  &   0.881  &   0.954  \\
        $\textbf{Ours}_\textbf{T}$ & M+S &   0.202  &   1.641  &   6.614  &   0.285  &   0.678  &   0.887  &   0.954  \\
        \bottomrule
    \end{tabular}
    \label{tab:kitti-frost}
\end{table}

\begin{table}[!thbp]
    \centering
    \renewcommand\arraystretch{0.8}
    \caption{Detailed performance comparison of monocular depth estimation on the KITTI Motion Blur test set. To ensure fairness in comparisons and avoid introducing any corrupted samples, all the models mentioned above are trained on the KITTI Clear training dataset. Upper Section: ResNet-based Architectures. Lower Section: Hybrid Architectures. $\downarrow$: Lower is better. $\uparrow$: Higher is better.}
    \begin{tabular}{cc|cccc|ccc}
    \toprule
    \multirow{2}{*}{Method} & \multirow{2}{*}{Knowledge} & \makecell{Abs\\Rel}$\downarrow$ & \makecell{Sq\\Rel}$\downarrow$ & \makecell{RMSE}$\downarrow$ & \makecell{RMSE\\Log}$\downarrow$ & $\delta_1\uparrow$ & $\delta_2\uparrow$ & $\delta_3\uparrow$ \\
    \midrule 
        MonoDepth2 & M &   0.242  &   2.209  &   7.613  &   0.343  &   0.613  &   0.833  &   0.926  \\
        SCDepthV1 & M &   0.334  &   3.530  &   8.972  &   0.462  &   0.492  &   0.728  &   0.848  \\
        HRDepth & M &   0.345  &   3.599  &   9.314  &   0.482  &   0.476  &   0.708  &   0.833  \\
        SCDepthV3 & M+D &   0.282  &   2.630  &   8.834  &   0.396  &   0.537  &   0.777  &   0.892  \\
        DynaDepth & M+I &   0.345  &   3.654  &   9.167  &   0.455  &   0.487  &   0.715  &   0.848  \\
        SGDepth & M+S &   0.259  &   2.305  &   7.886  &   0.371  &   0.578  &   0.811  &   0.909  \\
        PackNet-SfM & M+S &   0.291  &   2.725  &   8.649  &   0.402  &   0.519  &   0.770  &   0.894  \\
        $\textbf{Ours}_\textbf{R}$ & M+S &   0.238  &   2.087  &   7.447  &   0.338  &   0.617  &   0.840  &   0.931  \\
        \midrule
        LiteMono & M &   0.252  &   2.094  &   7.555  &   0.345  &   0.582  &   0.825  &   0.926  \\
        MonoViT & M &   0.189  &   1.442  &   6.651  &   0.275  &   0.699  &   0.896  &   0.961  \\
        $\textbf{Ours}_\textbf{T}$ & M+S &   0.183  &   1.416  &   6.645  &   0.272  &   0.710  &   0.899  &   0.961  \\
        \bottomrule
    \end{tabular}
    \label{tab:kitti-motion}
\end{table}

\begin{table}[!thbp]
    \centering
    \renewcommand\arraystretch{0.8}
    \caption{Detailed performance comparison of monocular depth estimation on the KITTI Clear test set. Upper Section: ResNet-based Architectures. Lower Section: Hybrid Architectures. $\downarrow$: Lower is better. $\uparrow$: Higher is better.}
    \begin{tabular}{cc|cccc|ccc}
    \toprule
    \multirow{2}{*}{Method} & \multirow{2}{*}{Knowledge} & \makecell{Abs\\Rel}$\downarrow$ & \makecell{Sq\\Rel}$\downarrow$ & \makecell{RMSE}$\downarrow$ & \makecell{RMSE\\Log}$\downarrow$ & $\delta_1\uparrow$ & $\delta_2\uparrow$ & $\delta_3\uparrow$ \\
    \midrule 
        MonoDepth2 & M &   0.117  &   0.957  &   4.953  &   0.195  &   0.875  &   0.958  &   0.980  \\
        SCDepthV1 & M &   0.118  &   0.870  &   4.964  &   0.195  &   0.860  &   0.957  &   0.982  \\
        HRDepth & M &   0.106  &   0.815  &   4.598  &   0.184  &   0.891  &   0.963  &   0.982  \\
        SCDepthV3 & M+D &   0.117  &   0.735  &   4.682  &   0.186  &   0.866  &   0.961  &   0.984  \\
        DynaDepth & M+I &   0.114  &   0.890  &   4.914  &   0.194  &   0.876  &   0.959  &   0.981  \\
        SGDepth & M+S &   0.117  &   0.930  &   4.897  &   0.195  &   0.874  &   0.958  &   0.980  \\
        PackNet-SfM & M+S &   0.116  &   0.872  &   4.970  &   0.198  &   0.868  &   0.956  &   0.980  \\
        $\textbf{Ours}_\textbf{R}$ & M+S &   0.118  &   0.872  &   4.869  &   0.196  &   0.869  &   0.958  &   0.981  \\
        \midrule
        LiteMono & M &   0.113  &   0.910  &   4.874  &   0.193  &   0.882  &   0.960  &   0.980  \\
        MonoViT & M &   0.100  &   0.729  &   4.410  &   0.176  &   0.898  &   0.967  &   0.984  \\
        $\textbf{Ours}_\textbf{T}$ & M+S &   0.105  &   0.796  &   4.515  &   0.181  &   0.894  &   0.964  &   0.982  \\
        \bottomrule
    \end{tabular}
    \label{tab:kitti-clear}
\end{table}

\begin{table}[!thbp]
    \centering
    \renewcommand\arraystretch{0.8}
    \caption{Detailed performance comparison of monocular depth estimation on the DENSE Fog test set. To ensure fairness in comparisons and avoid introducing any corrupted samples, all the models mentioned above are trained on the KITTI Clear training dataset. Upper Section: ResNet-based Architectures. Lower Section: Hybrid Architectures. $\downarrow$: Lower is better. $\uparrow$: Higher is better.}
    \begin{tabular}{cc|cccc|ccc}
    \toprule
    \multirow{2}{*}{Method} & \multirow{2}{*}{Knowledge} & \makecell{Abs\\Rel}$\downarrow$ & \makecell{Sq\\Rel}$\downarrow$ & \makecell{RMSE}$\downarrow$ & \makecell{RMSE\\Log}$\downarrow$ & $\delta_1\uparrow$ & $\delta_2\uparrow$ & $\delta_3\uparrow$ \\
    \midrule 
        MonoDepth2 & M &   0.266  &   1.484  &   4.865  &   0.374  &   0.526  &   0.804  &   0.906  \\
        SCDepthV1 & M &   0.269  &   1.600  &   5.125  &   0.410  &   0.539  &   0.751  &   0.866  \\
        HRDepth & M &   0.262  &   1.402  &   4.707  &   0.366  &   0.533  &   0.797  &   0.920  \\
        SCDepthV3 & M+D &   0.257  &   1.568  &   5.087  &   0.380  &   0.578  &   0.788  &   0.892  \\
        DynaDepth & M+I &   0.271  &   1.589  &   5.191  &   0.408  &   0.519  &   0.776  &   0.891  \\
        SGDepth & M+S &   0.260  &   1.478  &   4.967  &   0.379  &   0.548  &   0.801  &   0.909  \\
        PackNet-SfM & M+S &   0.254  &   1.346  &   4.697  &   0.360  &   0.532  &   0.812  &   0.918  \\
        $\textbf{Ours}_\textbf{R}$ & M+S &   0.240  &   1.262  &   4.567  &   0.342  &   0.580  &   0.834  &   0.924  \\
        \midrule
        MonoViT & M &   0.239  &   1.223  &   4.600  &   0.345  &   0.556  &   0.824  &   0.926  \\
        LiteMono & M &   0.325  &   1.922  &   4.863  &   0.383  &   0.422  &   0.727  &   0.926  \\
        $\textbf{Ours}_\textbf{T}$ & M+S &   0.224  &   1.032  &   3.907  &   0.312  &   0.583  &   0.848  &   0.948  \\
        \bottomrule
    \end{tabular}
    \label{tab:dense-fog}
\end{table}

\begin{table}[!thbp]
    \centering
    \renewcommand\arraystretch{0.8}
    \caption{Detailed performance comparison of monocular depth estimation on the DENSE Rain test set. To ensure fairness in comparisons and avoid introducing any corrupted samples, all the models mentioned above are trained on the KITTI Clear training dataset. Upper Section: ResNet-based Architectures. Lower Section: Hybrid Architectures. $\downarrow$: Lower is better. $\uparrow$: Higher is better.}
    \begin{tabular}{cc|cccc|ccc}
    \toprule
    \multirow{2}{*}{Method} & \multirow{2}{*}{Knowledge} & \makecell{Abs\\Rel}$\downarrow$ & \makecell{Sq\\Rel}$\downarrow$ & \makecell{RMSE}$\downarrow$ & \makecell{RMSE\\Log}$\downarrow$ & $\delta_1\uparrow$ & $\delta_2\uparrow$ & $\delta_3\uparrow$ \\
    \midrule 
        MonoDepth2 & M &   0.274  &   1.554  &   4.962  &   0.381  &   0.517  &   0.792  &   0.901  \\
        SCDepthV1 & M &   0.266  &   1.570  &   5.118  &   0.396  &   0.537  &   0.765  &   0.882  \\
        HRDepth & M &   0.262  &   1.394  &   4.730  &   0.366  &   0.528  &   0.804  &   0.918  \\
        SCDepthV3 & M+D &   0.253  &   1.513  &   5.087  &   0.377  &   0.581  &   0.797  &   0.894  \\
        DynaDepth & M+I &   0.281  &   1.657  &   5.311  &   0.415  &   0.504  &   0.771  &   0.886  \\
        SGDepth & M+S &   0.252  &   1.413  &   4.912  &   0.372  &   0.563  &   0.806  &   0.911  \\
        PackNet-SfM & M+S &   0.256  &   1.370  &   4.765  &   0.365  &   0.533  &   0.811  &   0.913  \\
        $\textbf{Ours}_\textbf{R}$ & M+S &   0.247  &   1.355  &   4.642  &   0.348  &   0.587  &   0.821  &   0.918  \\
        \midrule
        MonoViT & M &   0.237  &   1.280  &   4.767  &   0.354  &   0.564  &   0.822  &   0.914  \\
        LiteMono & M+S &   0.322  &   1.877  &   4.787  &   0.378  &   0.414  &   0.757  &   0.928  \\
        $\textbf{Ours}_\textbf{T}$ & M+S &   0.224  &   1.083  &   4.046  &   0.321  &   0.571  &   0.841  &   0.941  \\
        \bottomrule
    \end{tabular}
    \label{tab:dense-fog}
\end{table}

\section{Encoder Feature Visualization Techniques}

As displayed in Algorithm \ref{alg:pca-vis}, the visualization \(V_t\) emerges as a powerful tool to demystify the complex feature representations generated by the texture encoder and structure encoder in \(\textbf{Ours_T}\). By leveraging Principal Component Analysis (PCA) to distill the multidimensional feature space into a singular, interpretable dimension. This method effectively highlights the spatial distribution and intensity variations within the encoded features, offering insights into the encoder's focus and its differentiation of input data aspects. The transition from high-dimensional representations to a comprehensible visual format not only aids in qualitative analysis but also enhances our understanding of the encoder's operational nuances. Thus, \(V_t\) serves as a bridge, translating abstract numerical data into a visual narrative that succinctly conveys the encoder's learning dynamics and its interpretative power over the input data.

\begin{algorithm}
\caption{Visualization of Encoder-Generated Features}
\begin{algorithmic}
\label{alg:pca-vis}

\REQUIRE \(F_t\): The set of features generated from image \(I_t\) by an encoder, where \(F_t[c]\) represents the feature of channel \(c\).
\ENSURE \(V_t\): The visualization of the encoder's feature representations.

\STATE Initialize an empty list \({L}_\textrm{interp}\) to hold interpolated features for visualization.

\FOR{each channel \(c\) in \(F_t\)}
    \STATE Resize \(F_t[c]\) to a predefined shape using bilinear interpolation and add to \({L}_\textrm{interp}\).
\ENDFOR

\STATE \(U_t \leftarrow\) Concatenate all features in \({L}_\textrm{interp}\) along the feature dimension.
\STATE Flatten \(U_t\) spatially into a 2D array and transpose it, making features as rows.

\STATE Apply PCA to reduce the dimensionality of the transposed array to one dimension.

\STATE Compute the lower and upper bounds as the 5th and 95th percentiles of the PCA-reduced array, respectively.
\STATE Clip the PCA-reduced array within these bounds.
\STATE Normalize the normalized array to the range [0, 255] and convert it to an unsigned 8-bit gray-scale image.

\STATE Reshape the scaled array back to the original shape of \(I_t\) to obtain the visualization \(V_t\).
\end{algorithmic}
\end{algorithm}

\section{Detailed Feature Visualization Results}

\begin{figure}
    \centering
    \includegraphics[width=0.8\textwidth]{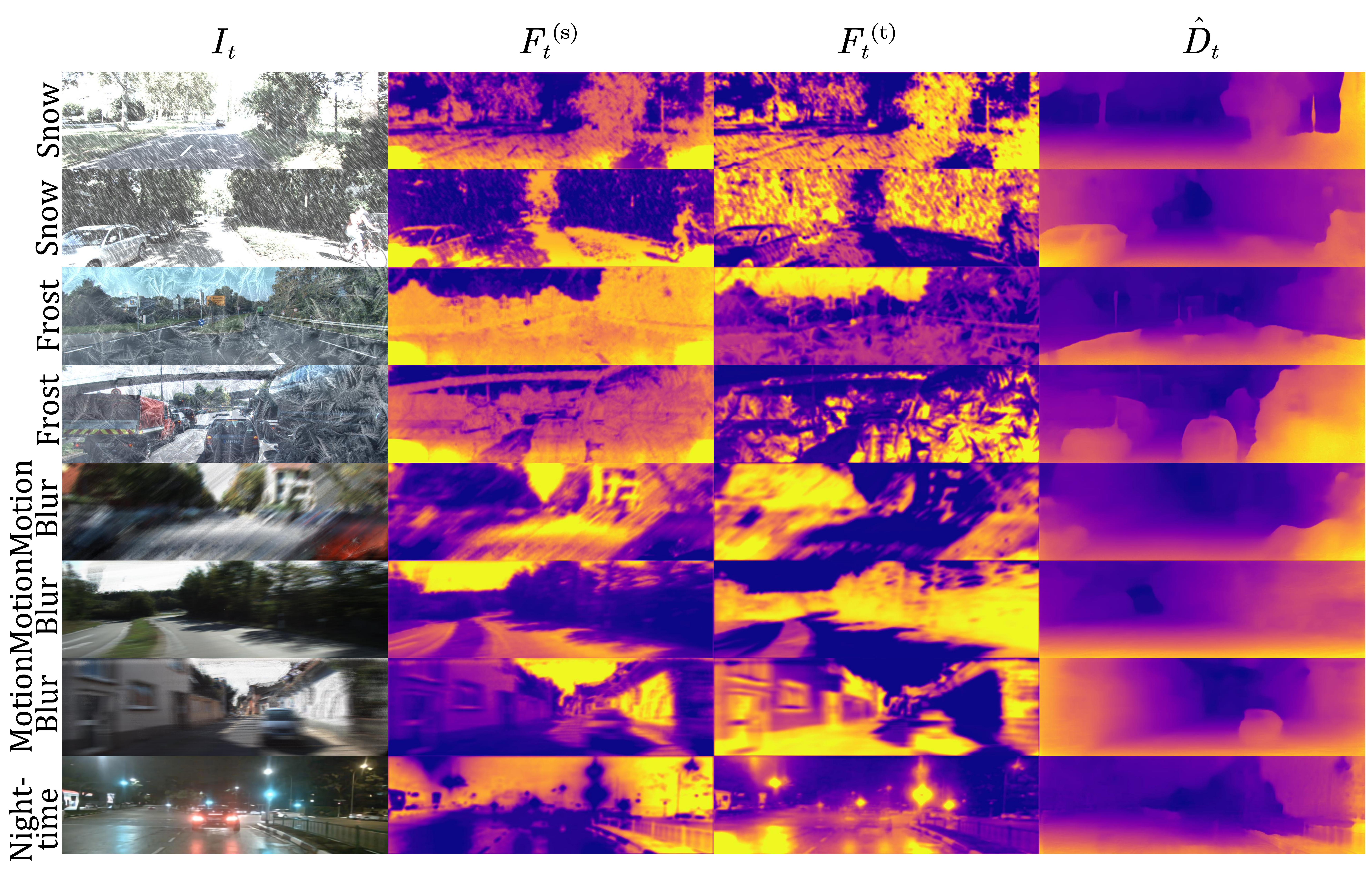}
    \caption{Detailed visualization of intermediate feature maps. For features $F_t^{(\mathrm{s})}$ and $F_t^{(\mathrm{t})}$, yellow indicates high values. For predicted depth $\hat{D}_t$, purple denotes high values. We provide a detailed description of the computation method for visual feature maps in the supplementary materials.}
    \label{fig:feature-detailed}
\end{figure}

We provide representative feature visualization results in the main text. In the supplementary material, we present feature visualizations in a broader range of scenarios to demonstrate the generalizability of our feature interpretability in Figure \ref{fig:feature-detailed}.

\section{Visualization of Depth}

\begin{figure*}[!t]
    \centering
    \includegraphics[width=0.8\textwidth]{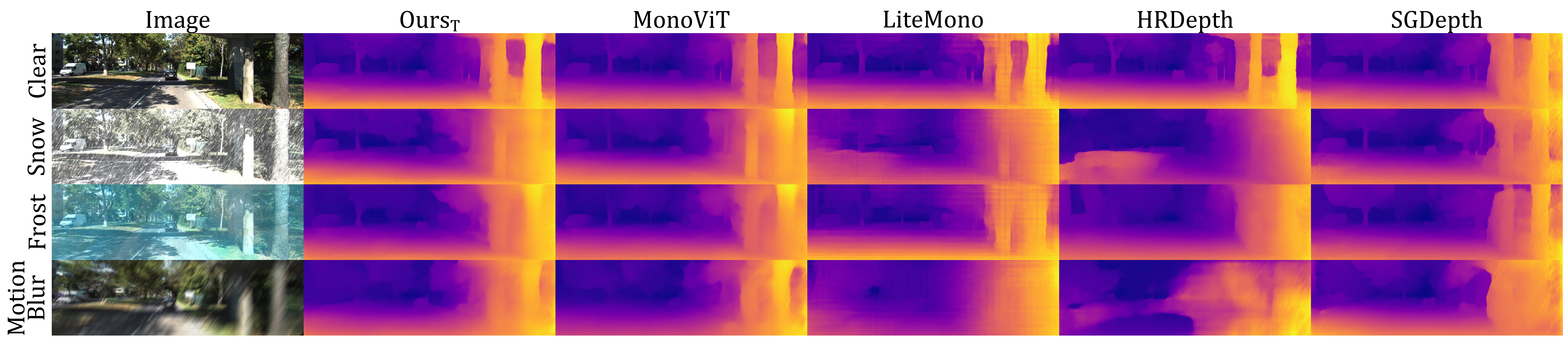}
    \caption{Visualized depth comparison of $\mathrm{Ours}_\mathrm{T}$, MonoViT, LiteMono, HRDepth, and SGDepth.}
    \label{fig:depth-comparison}
\end{figure*}

We visualize the estimated depth results on an image with a complex scene from the KITTI test sequence as displayed in Figure \ref{fig:depth-comparison}, using different models including MonoViT, LiteMono, HRDepth, and SGDepth. The visualizations show $\mathrm{Ours}_\mathrm{T}$ can identify the overhead bridge in dark areas for clear scenes, while experiencing less scene interference compared to other methods.


%
%